\documentclass[journal]{IEEEtran}
\usepackage{cite}
\usepackage{algorithmic}
\usepackage{algorithm}
\usepackage{multicol,multirow}
\usepackage{bm}
\usepackage{epstopdf}
\usepackage[figuresright]{rotating}
\usepackage{threeparttable}
\usepackage{makecell}
\usepackage{amsmath, subfigure, float}
\usepackage{eurosym}
\usepackage{mathrsfs}
\usepackage{fancyhdr,multicol}
\usepackage{threeparttable}
\usepackage{subfloat}
\usepackage{graphicx}
\usepackage{booktabs}
\usepackage{url}

\newcommand{\ie}{\textit{i.e.}}

\ifCLASSINFOpdf
\else
\fi
\begin{document}
		%
		\title{PipeOptim: Ensuring Effective 1F1B Schedule with Optimizer-Dependent Weight Prediction}
		%
		%
		%
		
		\author{Lei Guan,~\IEEEmembership{Member,~IEEE,}
			Dongsheng Li, Yongle Chen, Jiye Liang,~\IEEEmembership{Fellow,~IEEE,} Wenjian Wang,  Xicheng Lu
			\thanks{Lei Guan and Yongle Chen are with the College of Computer Science and Technology, Taiyuan University of Technology, Taiyuan 030024, China. E-mail: guanleimath@163.com, chenyongle@tyut.edu.cn.}
			\thanks{Dongsheng Li and Xicheng Lu are with the School of Computer, National University of Defense Technology, Changsha 410073, China. E-mail: dsli@nudt.edu.cn, xclu@nudt.edu.cn.}
			\thanks{Jiye Liang and Wenjian Wang are with the School of Computer and Information Technology, Shanxi University, Taiyuan 030006, China. E-mail: ljy@sxu.edu.cn, wjwang@sxu.edu.cn.}
			\thanks{Corresponding author: Lei Guan.}
			\thanks{Manuscript received April 19, 2005; revised August 26, 2015.}}
		
		%
		%

	\markboth{Journal of \LaTeX\ Class Files,~Vol.~14, No.~8, August~2015}%
	{Shell \MakeLowercase{\textit{et al.}}: Bare Demo of IEEEtran.cls for IEEE Journals}
	%



	\maketitle
	
	\begin{abstract}
		Asynchronous pipeline model parallelism with a ``1F1B'' (one forward, one backward) schedule generates little bubble overhead and always provides quite a high throughput. However, the ``1F1B'' schedule inevitably leads to weight inconsistency and weight staleness issues due to the cross-training of different mini-batches across GPUs. To simultaneously address these two problems, in this paper, we propose an optimizer-dependent weight prediction strategy (a.k.a PipeOptim) for asynchronous pipeline training. The key insight of our proposal is that we employ a weight prediction strategy in the forward pass to approximately ensure that each mini-batch uses consistent and staleness-free weights to compute the forward pass of the ``1F1B'' schedule. To be concrete, we first construct the weight prediction scheme based on the update rule of the used optimizer when training the deep neural network models. Then throughout the  ``1F1B'' pipeline training, each mini-batch is mandated to execute weight prediction, subsequently employing the predicted weights to perform the forward pass. As a result, PipeOptim 1) inherits the advantage of the ``1F1B'' schedule and generates high throughput, and 2) can ensure effective parameter learning regardless of the type of the used optimizer. We conducted extensive experimental evaluations using nine different deep-learning models to verify the effectiveness of our proposal. The experiment results demonstrate that PipeOptim outperforms the other five popular pipeline approaches including GPipe, PipeDream, PipeDream-2BW, SpecTrain, and XPipe. 
	\end{abstract}
	
	\begin{IEEEkeywords}
		pipeline model parallelism, deep neural network, weight prediction, asynchronous, learning efficiency.
	\end{IEEEkeywords}

	%
	\IEEEpeerreviewmaketitle

\section{Introduction}
\IEEEPARstart{D}{eep} learning has set milestones to progress toward human-level intelligence. In particular, for many machine learning tasks such as image and video analysis~\cite{carion2020end, chen2021pre}, natural language processing (NLP)~\cite{devlin2018bert,brown2020language}, and speech recognition~\cite{afouras2018deep, wang2020transformer}, many of the groundbreaking results were delivered through applying the deep neural networks (DNNs). Training a DNN model, however, is not trivial. The most popular approach for DNN training is data parallelism~\cite{goyal2017accurate, you2017scaling} where each accelerator (usually a GPU) holds the entire model parameters and is assigned with different sets of training data. However, data parallelism always suffers from excessive communication overhead due to the weight synchronization per iteration. Furthermore, data parallelism always hits another roadblock--it does not work once the DNN models with a huge number of parameters can not fit in a single GPU device. 
	
\begin{figure*}[ht]
		\centering
		\subfigure[Serial execution]{
			\label{fig:serial}
			\begin{minipage}[t]{0.55\textwidth}
				\centering
				\includegraphics[width=.95\linewidth]{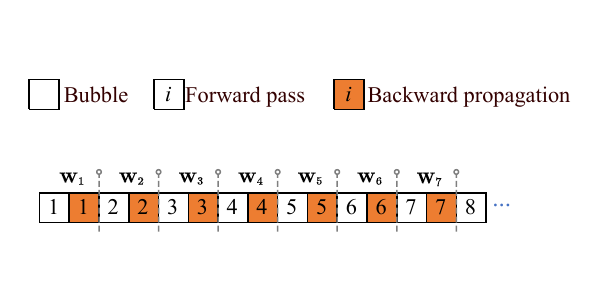}
			\end{minipage}
		} 
		\subfigure[GPipe]{
			\label{fig:gpipe}
			\begin{minipage}[t]{0.40\textwidth}
				\centering
				\includegraphics[width=.99\linewidth]{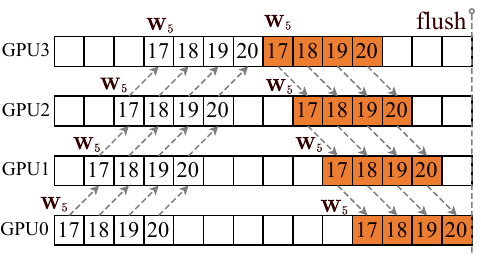}
			\end{minipage}
		}
		\subfigure[Naive approach]{
			\label{fig:naive}
			\begin{minipage}[t]{0.26\textwidth}
				\centering
				\includegraphics[width=.98\linewidth]{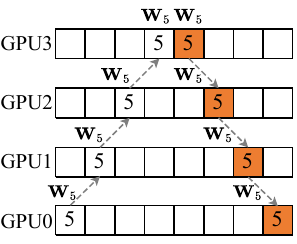}
			\end{minipage}
		}
		\subfigure[PipeDream]{
			\label{fig:pipedream}
			\begin{minipage}[t]{0.70\textwidth}
				\centering
				\includegraphics[width=.98\linewidth]{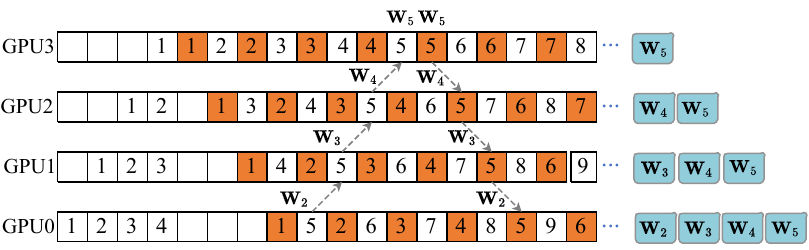}
			\end{minipage}
		}
		\\
		\caption{Timelines of serial execution, GPipe, the naive approach, and PipeDream. In Figures~\ref{fig:gpipe}, \ref{fig:naive}, and \ref{fig:pipedream}, the grey dashed arrows represent the pipeline training of the 5\emph{th} mini-batch (micro-batches 17, 18, 19, and 20 for GPipe). The blue squares on the right side of Figure~\ref{fig:pipedream} indicate the weights needed to be maintained during the training period of the 5\emph{th} mini-batch. The blue squares on the right side of Figure~\ref{fig:pipeoptim} show the maintained weights for the forward pass of the 5\emph{th} mini-batch.}
		\label{fig:pipeline}
	\end{figure*}

	
In recent years, pipeline model parallelism (PMP) has been attracting increasing attention and has become the most popular approach for training DNN models with large numbers of parameters. Synchronous PMP approaches, such as GPipe~\cite{huang2019gpipe}, often suffer from under-utilization of compute resources due to bubble overhead, as depicted in Figure~\ref{fig:gpipe}. The most popular asynchronous PMP approaches, such as PipeDream~\cite{narayanan2019pipedream} and PipeDream-2BW~\cite{narayanan2021memory}, address the bubble overhead problem by adopting the ``1F1B'' scheduling strategy and using the weight stashing technique to resolve the weight inconsistency issue. Yet, the weight stashing technique cannot handle the weight staleness issue which degrades both convergence and model accuracy. On the other hand, the weight prediction technique~\cite{guan2024xgrad,chen2018efficient} has been successfully applied to simultaneously address the weight inconsistency and staleness issues in asynchronous pipeline training. Notable examples include SpecTrain~\cite{chen2018efficient} and XPipe~\cite{guan2019xpipe}, whose performance, however, heavily relies on the optimizer used. 
	
To this end, we propose a novel and efficient PMP approach called PipeOptim. Like PipeDream and PipeDream-2BW, PipeOptim adopts the ``1F1B'' schedule to achieve high GPU resource utilization and throughput. Instead of using the weight stashing technique, PipeOptim introduces an optimizer-dependent weight prediction strategy based on the update rule of the used optimizer. Remarkably, PipeOptim achieves effective parameter learning from the optimizer's perspective, which distinguishes it from PipeDream and PipeDream-2BW, both of which rely on the intuitive weight stashing technique. As a result, PipeOptim can simultaneously address the weight inconsistency and weight staleness issues incurred by the ``1F1B'' schedule, effectively realizing the goals of high throughput and effective parameter learning, regardless of the type of the optimizer used. Furthermore, PipeOptim requires each GPU to maintain at most two versions of weights, striking a good balance among GPU utilization, convergence, and memory consumption, ultimately delivering high overall performance. 
	
We evaluated PipeOptim using nine different DNN models spanning four machine-learning tasks. The experimental results, presented in detail, demonstrate the effectiveness of our proposal. Thanks to the ``1F1B'' schedule, PipeOptim consistently achieves higher throughput than GPipe, regardless of the benchmark models used.  As opposed to PipeDream and PipeDream-2BW, PipeOptim effectively mitigates accuracy degradation, achieving model accuracy comparable to (or even slightly better than) GPipe. At the same time, The performance of PipeOptim is independent of the type of the optimizer used, outperforming XPipe and well addressing the limitation of SpecTrain, which performs well only when the SGD with momentum (SGDM)~\cite{qian1999momentum} optimizer is used.

The main contributions of this paper can be summarized as follows.
\begin{itemize}
\item[1)] We establish the foundation of weight prediction in the asynchronous pipeline schedule and propose an effective method for predicting the future weights by performing computations based on the ``1F1B'' schedule's pipeline structure and the update rule of the chosen optimizer.
		
\item[2)] We further introduce an optimizer-dependent weight prediction strategy to achieve effective parameter learning for the ``1F1B'' schedule. Unlike SpecTrain~\cite{chen2018efficient} and XPipe~\cite{guan2019xpipe}, the efficiency of PipeOptim does not rely on any specific optimization method. To the best of our knowledge, this is the first work that simultaneously alleviates the weight inconsistency and weight staleness issues caused by the ``1F1B'' schedule.
		
\item[3)] We conducted extensive experimental evaluations using nine different DNN models to validate the effectiveness of our proposal. The experimental results demonstrate that PipeOptim achieves better trade-offs among GPU utilization, convergence, and memory consumption than the  five other popular PMP approaches, delivering the best overall performance among all evaluated PMP approaches. For example, when training ResNet-101 to the target accuracy, PipeOptim yields a speedup of 2.01X, 1.04X, 1.17X, 1.30X, and 1.27X speedup over GPipe, PipeDream, PipeDream-2BW, SpecTrain, and XPipe, respectively.
	\end{itemize}
	
	\section{Challenges of ``1F1B'' Schedule}\label{sec:challenge}
	A DNN model is typically composed of multiple neural layers, and the training of the DNN model involves learning the parameters of each layer through an optimization method. 
	One remarkable characteristic of DNN training is that the weights are updated in an iterative, sequential, and continuous manner~\cite{guan2024xgrad}. Figure~\ref{fig:serial} illustrates the serial execution of DNN training where the weights are initialized with $\mathbf W_1$ and, in a step-by-step fashion, updated to $\mathbf W_2$, $\mathbf W_3$, and so on.  During each mini-batch training, the computing device (default is a GPU) applies the current version of the weights to both forward pass and backward propagation, subsequently updating the weights to a new version. 
	We note that the weight update shown in Figure~\ref{fig:serial} is performed on a single GPU. Any parallel training approaches that maintain consistency with this serial execution can achieve effective parameter learning.

	 For PMP, high GPU utilization facilitates fast iteration speeds, while effective parameter learning contributes to fewer training iterations---both factors jointly determining the training duration of deep learning models~\cite{guan2024advances}. Therefore, achieving high GPU utilization while ensuring effective parameter learning is crucial for attaining efficient and effective pipeline training.
	Synchronous PMP approaches, such as the naive approach  (as show in~\ref{fig:naive}) and GPipe (as shown in~\ref{fig:gpipe}), generally suffer from bubble overhead due to periodic flushes. To overcome the bubble problem, the ``1F1B'' schedule was proposed in PipeDream~\cite{narayanan2019pipedream} and later adopted by other asynchronous PMP approaches such as PipeDream-2BW~\cite{narayanan2021memory}, SpecTrain~\cite{chen2018efficient}, and XPipe~\cite{guan2019xpipe}.
	As shown in Figure~\ref{fig:pipedream}, the primary advantage of the ``1F1B'' schedule is its extremely small bubble ratio ($\approx0\%$), which results in high GPU utilization and, consequently, high throughput. However, the ``1F1B'' schedule also directly leads to weight inconsistency and weight staleness issues, which negatively affect the learning efficiency of the model parameters. 
	
	In the following, we illustrate these two issues using the pipeline training of the 5\emph{th} mini-batch shown in Figure~\ref{fig:pipedream}.
	\begin{itemize}
		\item[1)] \textbf{Weight Inconsistency}: As shown in Figure~\ref{fig:pipedream}, GPU 0 performs the forward pass of the 5\emph{th} mini-batch with weights ${\mathbf W}_2$. However, by the time GPU 0 is ready to perform the backward propagation, the stage weights have been updated three times, reaching $\mathbf W_5$ after the backward propagations of mini-batches 2, 3, and 4, leading to an inconsistency. Similarly, mini-batches processed on GPUs 1 and 2 also experience weight inconsistency.
		
		\item[2)] \textbf{Weight Staleness}:  The ``1F1B'' schedule leads to the 5\emph{th} mini-batch to use different versions of weights for the forward pass (\ie, $\mathbf W_2$ for GPU 0, $\mathbf W_3$ for GPU 1, $\mathbf W_4$ for GPU 2, and $\mathbf W_5$ for GPU 3). Comparing these versions to the weight updates shown in Figure~\ref{fig:serial}, we observe that only $\mathbf W_5$ is the up-to-date version, while $\mathbf W_2$, $\mathbf W_3$, and $\mathbf W_4$ are stale, as they are generated before the backward propagation of the 4\emph{th} mini-batch. The staleness issue can degrade convergence and reduce model accuracy compared to serial execution.
	\end{itemize}
	
	To address the weight inconsistency issue, PipeDream proposes using the weight stashing technique, where all GPUs, except the last one, are required to maintain one version of weights for each in-flight mini-batch. Although the weight stashing technique can ensure the consistency between the forward pass and backward propagation of each mini-batch, it incurs additional and unbalanced memory usage. As shown on the right side of Figure~\ref{fig:pipedream}, the GPUs are required to store up to $D$ versions of weights, where $D$ denotes the pipeline depth. Furthermore, the weight stashing technique does not help resolve the weight staleness issue introduced by the ``1F1B'' schedule. As illustrated in Figure~\ref{fig:pipedream}, when using weight stashing, the 5\emph{th} mini-batch will use the same version of weights for both forward pass and backward propagation, that is, $\mathbf W_2$ on GPU 0, $\mathbf W_3$ on GPU 1, $\mathbf W_4$ on GPU 2, and $\mathbf W_5$ on GPU 3. However, when comparing these versions of weights with the serial execution shown in Figure~\ref{fig:serial}, all GPUs, except the last one, are still using stale weights for DNN training.

   Likewise, PipeDream-2BW~\cite{narayanan2021memory} adopts the double-buffered weight updates (2BW) technique,  a variant of the weight stashing method to address the weight inconsistency problem caused by the ``1F1B'' schedule. The 2BW technique requires PipeDream-2BW to maintain two versions of weights per GPU, offering higher memory efficiency than PipeDream. Yet, as with PipeDream, PipeDream-2BW still fails to resolve the weight staleness problem. 
	
	To simultaneously alleviate the inconsistency and staleness issues incurred by the ``1F1B'' schedule, Chen et al.~\cite{chen2018efficient} propose an approach called SpecTrain. SpecTrain inherits the ``1F1B'' schedule used in PipeDream, enabling the cross-execution of multiple mini-batches to achieve high GPU utilization. Instead of using the weight stashing technique, SpecTrain employs a weight prediction strategy to achieve effective parameter learning.  Motivated by the observation that the smoothed gradients used in SGDM reflect the trend of weight updates, SpecTrain uses the smoothed gradients, multiplied by the calculated weights version difference, to predict weights that will be used in the future pipeline unit. By allowing each mini-batch to use the predicted weights for both the forward pass and backward propagation, SpecTrain can simultaneously address the inconsistency and staleness issues. However, SpecTrain has significant conditional limitations, as it heavily depends on the update rule of SGDM and only performs well when using the SGDM as the optimizer. For other popular optimizers such as Adam~\cite{kingma2014adam} and AdamW~\cite{loshchilov2017decoupled} whose weight update rules do not solely rely on the smoothed gradients, the SpecTrain approach is no longer applicable. Similarly, XPipe~\cite{guan2019xpipe} builds its weight prediction strategy based on the Adam optimizer's update rule. Although XPipe has a broader range of applications than SpecTrain, as confirmed by the experiments in this paper, XPipe does not always perform well when using non-Adam optimizers (such as SGDM).

	\section{The PipeOptim Approach}
	To simultaneously address the weight inconsistency and staleness issues caused by the ``1F1B'' schedule, we propose an optimizer-dependent weight prediction strategy, referred to as PipeOptim. This strategy is constructed based on the update rule of the optimizer used to train the DNN model.
	
	As noted in Section~\ref{sec:challenge}, the ``1F1B'' schedule cause all stages, except the last, to 1) use inconsistent weights to do forward pass and backward propagation, and 2) apply stale weights for the forward pass. The key insight of PipeOptim is that for any mini-batch with weights $\mathbf W_t$, this mini-batch uses the predicted weights $\hat{\mathbf W}_{t+s}$ to perform forward pass, where $\hat{\mathbf W}_{t+s}$ represents the approximation of the future weights used by the backward propagation, which is available after $s$ times of continuous weight updates. This ensures that each mini-batch on any GPU approximately uses consistent weights for both forward and backward computations, effectively mitigating the staleness problem inherent in the ``1F1B'' schedule.

 \begin{figure*}[ht]
		\centering
			\begin{minipage}[t]{0.7\textwidth}
				\centering
				\includegraphics[width=.95\linewidth]{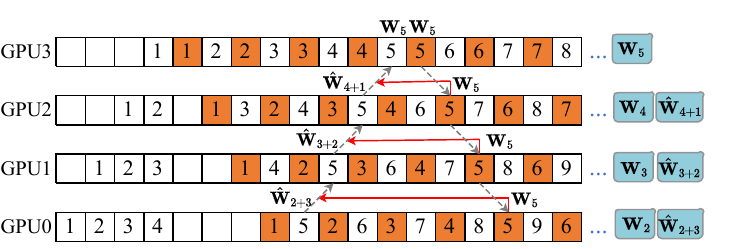}
			\end{minipage}
		\\
		\caption{Timelines of PipeOptim. The grey dashed arrows represent the pipeline training of the 5\emph{th} mini-batch; The blue squares on the right side of the figure illustrate the  weights maintained for the forward pass of the 5\emph{th} mini-batch.}
		\label{fig:pipeoptim}
	\end{figure*}
	
	Figure \ref{fig:pipeoptim} illustrates the main idea of PipeOptim on a 4-GPU computing system, using the pipeline training of the 5\emph{th} mini-batch as an example. The red arrowed lines represent the weight prediction performed by the 5\emph{th} mini-batch on each GPU. Each red arrowed line starts from the pipeline unit where the 5\emph{th} mini-batch performs the backward propagation and points to the pipeline unit where the 5\emph{th}  mini-batch is ready for the forward pass. On each GPU, except GPU 3, the 5\emph{th} mini-batch consistently uses the predicted weights (\ie, $\hat{\mathbf W}_{2+3}$ on GPU 0, $\hat{\mathbf W}_{3+2}$ on GPU 1, and $\hat{\mathbf W}_{4+1}$ on GPU 2) for the forward pass. Similarly, throughout the pipeline training, for any mini-batch with weights $\mathbf W_t$, the predicted weights $\hat{\mathbf W}_{t+s}$ are required ahead of the forward pass. Notably, in PipeOptim, only the forward pass of each mini-batch uses the predicted weights, while the backward propagation is performed normally using the available weights. As shown in Figure~\ref{fig:pipeoptim}, the backward propagation of the 5\emph{th} mini-batch directly makes use of $\mathbf W_5$, the staleness-free version of weights generated after the backward propagation of the $4$\emph{th} mini-batch.

	\subsection{Weight prediction by PipeOptim}
	
	In the following, we discuss how to bridge the gap between $\mathbf W_t$ and $\hat{\mathbf W}_{t+s}$ given the type of the optimization method and the available variables.
	
	
	\subsubsection{Weight prediction formula}
	Motivated by the weight prediction strategy proposed in XGrad~\cite{guan2024xgrad}, we derive the weight prediction formula for PipeOptim in the context of asynchronous pipeline training. The key insight of PipeOptim is based on the observation that on each GPU, the weights of a stage are updated continuously, and the relative variation of the weights reflects the trend of weight updates. 
	
	We begin by reconsidering the weight updates in serial execution, as shown in Figure~\ref{fig:serial}. We assume that the weights of any $t$-\emph{th} (where $t\geq 1$) mini-batch is ${\mathbf W}_t$ and the learning rate is $lr$. Over the next $s$ (where $s\geq 1$) times of mini-batch training, the DNN weights are updated as follows:
	\begin{align}
		&\mathbf W_{t+1}= \mathbf W_t - lr\cdot\Delta \mathbf{W}_{t}, \nonumber \\
		&\mathbf W_{t+2}= \mathbf W_{t+1} - lr\cdot\Delta \mathbf{W}_{t+1},
		\nonumber\\
		& \cdots \nonumber
		\\
		&\mathbf W_{t+s}= \mathbf W_{t+s-1} - lr\cdot\Delta \mathbf{W}_{t+s-1} .
		\label{W_1}
	\end{align}
	
	When summing all the weight update equations shown in~\eqref{W_1}, we can immediately get
	\begin{equation}
		\mathbf W_{t+s} = \mathbf W_{t} - lr \cdot \sum_{i=t}^{t+s-1} \Delta \mathbf{W}_i,
		\label{w_step}
	\end{equation}
	where $\Delta \mathbf{W}_i$ represents the relative increments of $ \mathbf W_{i+1}$ over $ \mathbf W_{i}$. Equation~\eqref{w_step} shows that, given the initial weights $\mathbf W_t$, $\mathbf W_{t+s}$ can be computed by letting $\mathbf W_t$ subtract the result of the learning rate timing the sum of $s$ continuous relative variation of the weights. Note that each relative variation of the weights (\ie, $\Delta \mathbf W_{i}$) should reflect the ``correct'' direction for updating the weights~\cite{guan2024xgrad}. Therefore, one can replace $\sum_{i=t}^{t+s-1} \Delta \mathbf{W}_i$ with $s\cdot \Delta \mathbf{W}_t$ in an effort to approximately  compute $\mathbf W_{t+s}$ for the case when only $\mathbf W_t$, $lr$, and $\Delta \mathbf{W}_{t}$ are available. 
	
	The discussions above directly generate the following weight prediction formula:
	\begin{equation}
		\hat{\mathbf W}_{t+s} \approx \mathbf W_t - lr \cdot s \cdot\Delta \mathbf{W}_{t}, 
		\label{weight_update}
	\end{equation}
	where $\mathbf W_t$ represent the given stage weights,  $\hat{{\mathbf W}}_{t+s}$ denotes the predicted weights for ${\mathbf W}_t$. For the weight prediction of PipeOptim, the parameter $s$ is always known as weight version differences which refers to the number of weight updates performed between the current pipeline unit and the future pipeline unit at which the mini-batch with index $t+s$ is ready to do backward propagation. 
	
	Given that  ${\mathbf W}_t$ and $lr$ are available, our focus then turns to how to compute $s$ and $\Delta {\mathbf W}_{t}$ to predict the future weights $\hat{\mathbf W}_{t+s}$ when using Equation~\eqref{weight_update} as the weight prediction formula. Notably, Equation~\eqref{weight_update} can be reduced to the weight prediction formula of XGrad~\cite{guan2024xgrad}. 
	

	\subsubsection{Computation of $s$}
	As shown in Figure~\ref{fig:pipeoptim}, the version difference $s$ depends heavily on the number of stages (\ie, the pipeline depth) and the index of each stage. PipeOptim lets each GPU compute $s$ via
	\begin{equation}
		s= D -rank -1, 
		\label{step}
	\end{equation}
	where $D$ refers to the pipeline depth and $rank$ is the index of a stage with $rank \in \{0, 1, \dots, size-1\}$.
	
	
	\subsubsection{Computation of $\Delta {\mathbf W}_{t}$}
	In Figure~\ref{fig:serial}, we assume that the $t$-\emph{th} mini-batch computes the gradients as $\mathbf{g}_{t} = \nabla f( \mathbf{W}_t)$, where $f(\cdot)$ is the loss function. According to the conclusions drawn in XGrad~\cite{guan2024xgrad}, given the gradients of the stochastic objective corresponding to the $t$-\emph{th} mini-batch, one can easily compute $\Delta \mathbf W_t$ according to the update rule of the used optimizer~\cite{guan2024xgrad}. 
	Table~\ref{tab:delta_theta} provides detailed information on how to compute $\Delta {\mathbf W}_{t}$ for the three most commonly used optimizers in deep learning: SGDM~\cite{qian1999momentum}, Adam~\cite{kingma2014adam}, and AdamW~\cite{loshchilov2017decoupled}.  We note that for SGDM, $u$ denotes the momentum factor and $\tau$ refers to the dampening for momentum. 
	For Adam and AdamW, $\mathbf m_t$ denotes the biased first-moment estimate;  $\mathbf v_t$ represents the biased second raw moment estimate; $\hat{\mathbf m}_{t}$ is the bias-corrected first-moment estimate; $\hat{\mathbf v}_{t}$ is the bias-corrected second raw moment estimates; ${\mathbf g}_t^2$ refers to element-wise square of the gradients, \ie, $\mathbf g_t^2=\mathbf g_t\odot \mathbf g_t$; $\epsilon$, $\gamma$ and $\lambda$ are constant values. 
	
	To summarize, during the ``1F1B'' schedule shown in Figure~\ref{fig:pipeoptim}, for any mini-batch with weights $\mathbf W_t$, PipeOptim lets each GPU perform the weight prediction ahead of the forward pass via
	\begin{equation}
		\hat{\mathbf W}_{t+s} = \mathbf W_t - lr \cdot s \cdot\Delta \mathbf{W}_{t}, 
		\label{eq:weight_update}
	\end{equation}
	where $lr$ is the learning rate, $s$ denotes the weight version difference, and $\Delta \mathbf{W}_{t}$ are computed based on the update rule of the used optimizer. The calculations of $\Delta \mathbf{W}_{t}$ for SGDM, Adam, and AdamW are shown in Table~\ref{tab:delta_theta}. PipeOptim utilizes both the gradients and cached optimizer states of the previous iteration step to calculate $\Delta \mathbf{W}_{t}$, where the gradients ${\mathbf g}_t$ are generated during the per-iteration backward propagation. We exemplify this with the SGDM optimizer as shown in Table~\ref{tab:delta_theta}. At the $t$-th iteration, PipeOptim uses the newly generated gradients ${\mathbf g}_t$ and the cached momentum variable ${\mathbf v}_{t-1}$ to compute ${\mathbf v}_t$, which is then returned as part of $\Delta \mathbf{W}_{t}$. For Adam and AdamW, $\Delta \mathbf{W}_{t}$ is calculated through the cached first-order and second-order moment (\ie, ${\mathbf m}_{t-1}$ and ${\mathbf v}_{t-1}$) along with the gradients ${\mathbf g}_t$.
	
	
	\begin{table}[ht]
		\centering
		\caption{The calculation of $\Delta {\mathbf W}_{t}$ for SGDM, Adam, and AdamW. }
		\begin{threeparttable} 
			\begin{tabular}{{cc}} 
				\toprule
				Optimizer & $\Delta {\mathbf W}_{t}$\\
				\midrule
				\makecell{SGDM} & \makecell{ $\begin{array}{ll}
						\quad \quad {\Delta \mathbf W}_{t} = \mathbf{m}_{t},\\  \textnormal{s.t.} \quad 
							{\mathbf m}_{t}= u\cdot \mathbf m_{t-1} + (1-\tau)\cdot \mathbf{g}_{t}.
						\end{array}$} \\
						\midrule
						Adam& \makecell{$\begin{array}{ll}
								\quad\quad	{\Delta \mathbf W}_{t} = \frac{\hat{\mathbf m}_{t}}{\sqrt{\hat{\mathbf v}_{t} } + \epsilon}, \\ \textnormal{s.t.} \ \left\{\begin{array}{ll}
									\mathbf{m}_{t} = \beta_1 \cdot \mathbf{m}_{t-1} + (1-\beta_1)\cdot \mathbf{g}_{t}, \\
									{\mathbf v}_{t}= \beta_2\cdot \mathbf v_{t-1} + (1-\beta_2) \cdot \mathbf{g}_{t}^2,\\
									\hat{\mathbf m}_{t}= \frac{\mathbf m_{t}}{1-\beta_1^{t}}, \\
									\hat{\mathbf v}_{t}= \frac{\mathbf v_{t}}{1-\beta_2^{t}}.\\
								\end{array}\right.
							\end{array}$}  \\
						\midrule
						AdamW & \makecell{$\begin{array}{ll}
								\quad\quad	{\Delta \mathbf W}_{t} = \frac{\hat{\mathbf m}_{t}}{\sqrt{\hat{\mathbf v}_{t} }+ \epsilon}, \\ \textnormal{s.t.} \ 	\left\{\begin{array}{ll}
									{\mathbf m}_{t}= \beta_1\cdot \mathbf m_{t-1} + (1-\beta_1) \cdot \mathbf{g}_{t},\\
									\mathbf{v}_{t} = \beta_2\cdot \mathbf{v}_{t-1} + (1-\beta_2)\cdot \mathbf{g}_{t}^2, \\
									\hat{\mathbf m}_{t}= \frac{\mathbf m_{t}}{1-\beta_1^{t}},\\
									\hat{\mathbf v}_{t}= \frac{\mathbf v_{t}}{1-\beta_2^{t}}. \\
								\end{array}\right.
							\end{array}$}   \\
						\bottomrule
					\end{tabular}
				\end{threeparttable}
				\label{tab:delta_theta}
			\end{table}

			\subsection{Comparision with asynchronous PMP approaches}
			In this section, we compare PipeOptim with other popular asynchronous PMP approaches including PipeDream, Pipedream-2BW, SpecTrain, and XPipe. Table~\ref{tab:comp} summarizes the detailed features of each approach.
			\begin{table*}[h!]
				\caption{Comparisons of typical pipeline parallelism approaches.}
				\begin{threeparttable} 
					\setlength{\tabcolsep}{2.6mm}
					\begin{tabular*}{\textwidth}{cccccc}
						\toprule
						Approaches& \makecell{Work Schedule}   & \makecell{Effective  Learning} & \makecell{Inconsistency Issue} & \makecell{Staleness Issue }    & \makecell{Weights  Memory$^*$ } \\
						\midrule
						PipeDream  & ``1F1B''  & weight stashing &    solved  &   unsolved &  [$M_\theta$, $D*M_\theta$]  \\
						PipeDream-2BW&  ``1F1B''  & weight stashing &  solved &  unsolved& $2*M_{\theta}$  \\
						SpecTrain & ``1F1B''  &SGDM-based weight prediction     &   \makecell{approximately  and \\ conditionally  solved} $^\dagger$ &  \makecell{approximately  and \\ conditionally solved} $^\dagger$& $2*M_{\theta}$ \\
                        XPipe & ``1F1B'' & Adam-based weight prediction & \makecell{approximately and \\ limitedly solved}$^\ddagger$   & \makecell{approximately and \\ limitedly  solved}$^\ddagger$  & $2*M_{\theta}$  \\
						PipeOptim & ``1F1B''  & optimizer-dependent weight prediction  &   approximately solved  &  approximately solved & [$M_\theta, 2*M_\theta$]\\
						\bottomrule
					\end{tabular*}
					\begin{tablenotes}    
						\footnotesize               
						\item[*] $M_{\theta}$ denotes the memory consumption required to store a stage.     
						\item[$\dagger$] SpecTrain does not work when using non-SGDM optimizers. 
						\item[$\ddagger$] The performance of XPipe with non-Adam optimizers depends on the evaluated DNN models. 
					\end{tablenotes}  
				\end{threeparttable}
				\label{tab:comp}
			\end{table*}
			
			\subsubsection{PipeOptim vs. PipeDream \& PipeDream-2BW}
			All the approaches adopt the ``1F1B'' schedule but employ different strategies to ensure effective parameter learning. PipeDream and PipeDream-2BW use the weight stashing technique, while PipeOptim utilizes an optimizer-dependent weight prediction strategy. The weight stashing technique guarantees that each mini-batch uses consistent weights for both the forward pass and backward propagation. However, it leaves the staleness issue unsolved. In contrast, the optimizer-dependent weight prediction strategy adopted by PipeOptim can address both the inconsistency and staleness issues incurred by the ``1F1B'' schedule simultaneously. Furthermore, the weight stashing technique requires more memory for storing weights$\textendash$ up to $D$ versions of weights for PipeDream and 2 versions of weights for PipeDream-2BW. In contrast, PipeOptim requires each GPU to maintain only one or two versions of weights, making it more memory efficient.
			
			\subsubsection{PipeOptim vs. SpecTrain and XPipe}
			All approaches adopt the ``1F1B'' schedule and employ weight prediction technique to ensure effective parameter learning. The weight prediction formulas of SpecTrain, XPipe, and PipeOptim are summarized as follows.
			\begin{equation}
				\begin{aligned}
					&\text{SpecTrain:} \ \  \hat{\mathbf W}_{t+s} = \mathbf W_t - lr \cdot s \cdot \mathbf{m}_{t-1}, \\
                    &\text{XPipe:} \ \  \hat{\mathbf W}_{t+s} = \mathbf W_t - lr \cdot s \cdot \frac{\bar{\mathbf{v}}_{t-1}}{\sqrt{\bar{\mathbf{m}}_{t-1} + \epsilon}}, \\
					&\text{PipeOptim:} \ \  \hat{\mathbf W}_{t+s} = \mathbf W_t - lr \cdot s \cdot\Delta \mathbf{W}_{t},
				\end{aligned}
				\label{eq:comparison}
			\end{equation}
		where $\mathbf{m}_{t-1}$ refers to the smoothed gradients in SGDM; $\bar{\mathbf{m}}_{t-1}$ is the bias-corrected first-moment estimate, $\bar{\mathbf{v}}_{t-1}$ is the bias-corrected second-moment estimate in Adam; and $\Delta\mathbf{W}_{t}$ is computed dynamically according to the update rule of the optimizer being used.
			
		Equation~\eqref{eq:comparison} shows that the weight prediction schemes of SpecTrain and XPipe are built dedicated based on the update rule of SGDM and Adam, respectively. In contrast, the weight prediction scheme of PipeOptim dynamically adapts based on the optimizer used to train the DNN model. For SpecTrain, the weight inconsistency and staleness issues are conditionally solved because SpecTrain does not work once using non-SGDM optimizers. XPipe, on the other hand, uses an Adam-based weight prediction strategy for all optimizers, which limits its performance with non-Adam optimizers. In contrast to SpecTrain and XPipe, the weight prediction formula in PipeOptim adjusts dynamically according to the chosen optimizer, ensuring its effectiveness regardless of the optimizer used. Additionally, while both SpecTrain and XPipe require weight prediction to be performed twice$\textendash$once during the forward pass and once during the backward propagation$\textendash$PipeOptim only requires a single weight prediction during the forward pass. This results in a lower computational cost compared to both SpecTrain and XPipe.
        

			\section{The PipeOptim Workflow}~\label{sec:algorithm}
			Algorithm \ref{alg1} describes the workflow of PipeOptim on a multi-GPU computing node. By default, each GPU holds a single stage, takes charge of updating its parameters, and meets the one-to-one correspondence. During the pipeline training, each GPU works in parallel and executes the same workflow outlined in Algorithm~\ref{alg1}.
			
			Before the pipeline training starts, each GPU initializes the stage weights or loads the stage weights before the pipeline starts (line 1). The weight version index is then set to $t\leftarrow 1$ (line 2).
			Throughout the pipeline training, each GPU runs a loop until the mini-batch training is complete (lines 3 to 16). At each iteration, the forward pass and backward propagation are executed in turn. Note that weight prediction is performed only by the GPUs with $rank<D-1$ (lines 4-10). The last GPU directly uses ${\mathbf W}_t$ for both forward pass and backward propagation (lines 12 and 14). On the front $D-1$ GPUs, PipeOptim always lets each mini-batch use the predicted future weights, $\hat{\mathbf W}_{t+s}$, to compute the forward pass. Specifically, on each GPU, weight prediction always goes ahead of the forward pass, with the currently available weights being cached first (line 5). The weight version difference is then calculated using Equation~\eqref{step} (line 6), and $\Delta {\mathbf W}_t$ is computed by a quick lookup of Table~\ref{tab:delta_theta} (line 7). Following that, PipeOptim applies Equation~\eqref{eq:weight_update} to compute the predicted weights $\hat{\mathbf W}_{t+s}$ (line 8). After the mini-batch computes the forward pass using the predicted weights, the cached weights are restored and used for backpropagation by another mini-batch.
			\begin{algorithm}[htb]
				\centering
				\caption{The workflow of PipeOptim}
				\label{alg1}
				\begin{algorithmic}[1]
					\REQUIRE $rank$, $D$, and $lr$.
					\STATE {Initialize or load the stage weights ${\mathbf W}$.}
					\STATE{$t\leftarrow 1$.}
					\WHILE {Mini-batch remains}
					\IF{$rank < D-1$}
					\STATE{Cache the current weight $\mathbf W_t$.}
					\STATE{Compute version difference $s$ using Eq.~\eqref{step}.}
					\STATE{Compute $\Delta \mathbf{W}_{t}$ using the formula shown in Table~\ref{tab:delta_theta}.}
					\STATE{Compute the predicted weights $\hat{\mathbf W}_t$ using Eq.~\eqref{eq:weight_update}.}
					\STATE {Do forward pass using $\hat{\mathbf W}_{t+s}$.}
					\STATE{Recover the cached weights $\mathbf W_t$. }
					\ELSE
					\STATE{Do forward pass using $\mathbf W_{t}$.}
					\ENDIF
					\STATE {Do backward propagation using $\mathbf W_{t}$.}
					\STATE{$t\leftarrow t+1$}
					\ENDWHILE
					\ENSURE{${\mathbf W}$}	
				\end{algorithmic}
			\end{algorithm}
		
		  Note that PipeOptim can be easily implemented on top of the popular pipeline parallelism frameworks such as PipeDream~\cite{narayanan2019pipedream} and GPipe~\cite{huang2019gpipe}. We do not report the implementation details, as they are beyond the main contributions of this paper. The source code of PipeOptim is available at: \url{https://github.com/guanleics/PipeOptim}.

	\section{Experimental Results}
	\subsection{Experimental Setup}
	We conducted the experiments on a multi-GPU computing platform consisting of three computing nodes, each equipped with four NVIDIA Tesla P100 GPUs (16 GB of device memory each), and powered by an Intel Xeon E5-2680 CPU operating at 2.40 GHz.

			
			
\textbf{Datasets.} We conducted experiments using four machine-learning tasks and five datasets: 1) Image classification with the CIFAR-100 and the Tiny-ImageNet~\cite{yao2015tiny} datasets; 2) Machine translation with the WMT16 English to German  (WMT En $\rightarrow$ De) dataset~\cite{sennrich2016edinburgh}; 3) Sentiment analysis with the IMDb Movie Review Sentiment Dataset~\cite{maas2011learning}. 4) Natural language processing (NLP) with the Wikipedia dataset. The CIFAR-100 dataset consists of 60,000 32$\times$32 color images. All the images are grouped into 100 classes with 500 training images and 100 testing images per class. Meanwhile, each CIFAR image was normalized with mean=[0.491,0.482,0.447] and std=[0.202, 0.199, 0.201]. The Tiny-ImageNet dataset contains 200 classes each having 500 training images and 50 validation images.  Each $64\times64\time3$ image was first scaled up to $224\times224\times3$, subsequently being normalized with mean=[0.485, 0.456, 0.406] and std=[0.229, 0.224, 0.225]. The IMDb Movie Review Sentiment Dataset encompasses a collection of 25,000 training movie reviews and an additional 25,000 testing reviews. Each review within the dataset has been labeled with a sentiment classification, specifically categorized as either positive or negative. The WMT dataset uses the WMT16 English to German dataset, which contains 36 million sentence pairs for training and the ``newstest2014'' dataset, consisting of 2,999 sentence pairs for validation.

\textbf{Models.} We selected nine different deep learning models as the benchmark DNN models, spanning across four different applications: 1) AlexNet~\cite{krizhevsky2012imagenet}, 2) VGG-16~\cite{simonyan2014very}, 3) ResNet-101~\cite{he2016deep}, 4) GoogleNet~\cite{szegedy2015going}, 5) Inception-V3~\cite{Szegedy_2016_CVPR}, 6) Residual LSTM~\cite{kim2017residual}, 7)  Bert-48~\cite{devlin2018bert}, 8) Google Neural Machine Translation (GNMT)~\cite{wu2016google} with 8 LSTM layers (dubbed as GNMT-8), and 9) GNMT with 16 LSTM layers (dubbed as GNMT-16).
			
We used the three most popular optimizers including SGDM~\cite{qian1999momentum}, Adam~\cite{kingma2014adam}, and AdamW~\cite{loshchilov2017decoupled} to optimize the DNN weights. In particular, we used SGDM and AdamW for the image classification task, Adam and AdamW for the sentiment analysis task, and Adam for the machine translation and NLP tasks. Each optimizer was set with the default hyper-parameters. To be concrete, for SGDM, we set momentum with 0.9 and weight decay with $5e^{-4}$. For both Adam and AdamW, we set $\beta_1=0.9$ and $\beta_2=0.999$ .

\textbf{Evaluating Approaches.}
In the experiments, we mainly compared PipeOptim with five other representative PMP approaches: GPipe~\cite{huang2019gpipe}, PipeDream~\cite{narayanan2019pipedream}, PipeDream-2BW~\cite{narayanan2021memory}, SpecTrain~\cite{chen2018efficient}, and XPipe~\cite{guan2019xpipe}. For PipeDream and PipeDream-2BW, we used the source code released on the GitHub\footnote{\url{https://github.com/msr-fiddle/pipedream}}. Similarly, the code of SpecTrain is also publicly available online\footnote{\url{https://github.com/ntueclab/SpecTrain-PyTorch}}. 
To ensure a fair comparison, we implemented PipeOptim,  GPipe~\cite{huang2019gpipe}, and XPipe~\cite{guan2019xpipe} on top of the PipeDream implementation, which isolates the impact of different system implementations on performance. 
Additionally, the following two measures were taken to ensure a fair comparison. First, unless otherwise noted, all the evaluated approaches adopted the model partitioning strategy proposed in PipeDream~\cite{narayanan2019pipedream} to divide the DNN models into four stages, with each stage assigned to a separate GPU. Second, each approach employed the recomputation technique~\cite{huang2019gpipe} to reduce the memory consumption of storing intermediate activation, thereby improving GPU memory utilization. 
			

\subsection{Accuracy}\label{subsec:exp-convergence}
In this subsection, we make a comparison in terms of the obtained model accuracy. Since GPipe automatically reduces to the naive PMP approach when the mini-batch size equals the micro-batch size, we trained GPipe with $T=1$ where $T$ denotes the number of micro-batches in a mini-batch, to simulate the behavior of the naive approach and isolate the effects of model partition. Additionally, we consider the learning results of GPipe with $T=1$ as the baseline, because GPipe is a synchronous PMP approach and does not incur any model accuracy drop. To verify the effectiveness of PipeOptim and its robustness with respect to the used optimizers, we divide the experiments into three groups based on the optimizer used. Table~\ref{tab:group} summarizes the optimizers, models, and datasets used in each experiment group. In the first group, we used SGDM to optimize the CNN models including AlexNet, ResNet-101, and Inception-V3. In the second group, we used AdamW to optimize VGG-16, GoogleNet, and Residual LSTM. In the third group, we used Adam to optimize Residual LSTM, GNMT-8, and GNMT-16.
			
			\begin{table}[htbp]
				\caption{Summarization of experimental setting of each group.}
				\begin{center}
					\begin{tabular}{cccc}
						\toprule
						Groups & Optimizer& Models& Dataset  \\
						\midrule
						Group-1 & SGDM & \makecell{AlexNet,  ResNet-101, \\Inception-V3 }& Tiny-Imagenet \\
						\cline{3-4}
						\multirow{2}{*}{Group-2}& \multirow{2}{*}{AdamW }&  VGG-16, GoogleNet& CIFAR-100  \\
						& &  Residual LSTM& IMDb \\
						\cline{3-4}
						\multirow{2}{*}{Group-3} & \multirow{2}{*}{Adam} 
						&  Residual LSTM & IMDb  \\
						&  &  GNMT-8, GNMT-16 & WMT-16 En$\rightarrow$De \\
						\bottomrule
					\end{tabular}
					\label{tab:group}
				\end{center}
			\end{table}
			
			In Group-1, we used SGDM with a momentum factor $\gamma=0.9$ and a weight decay of 5e-4. We trained AlexNet on Tiny-ImageNet for 70 epochs with a mini-batch size of 128, and trained ResNet-101 and Inception-V3 with a mini-batch size of 64. The learning rate was initially set to 0.01 and reduced by a factor of 10 at the 40\emph{th} and 60\emph{th} epochs. In Group-2, we used AdamW to optimize VGG-16 and GoogleNet on CIFAR-100 for 120 epochs with a mini-batch size of 128. The learning rate was initialized to 0.001 and decreased by a factor of 10 at the 90\emph{th} epoch. Additionally we used AdamW to optimize Residual LSTM for 50 epochs with a mini-batch size of 256. In Group-3, we used Adam to optimize Residual LSTM for 50 epochs with a mini-batch size of 256. The initial learning rate was 0.001 and decayed with the polynomial scheduling strategy~\cite{mishra2019polynomial}. In addition, we also trained GNMT-8 and GNMT-16 on the WMT En $\rightarrow$ De dataset using the Adam optimizer. We trained both GNMT-8 and GNMT-16 for 8 epochs with a mini-batch size of 64 and a fixed learning rate of $3e^{-4}$. 
			
\begin{figure*}[h!]
				\centering
				\subfigure[AlexNet on Tiny-ImageNet]{\includegraphics[width=0.32\textwidth]{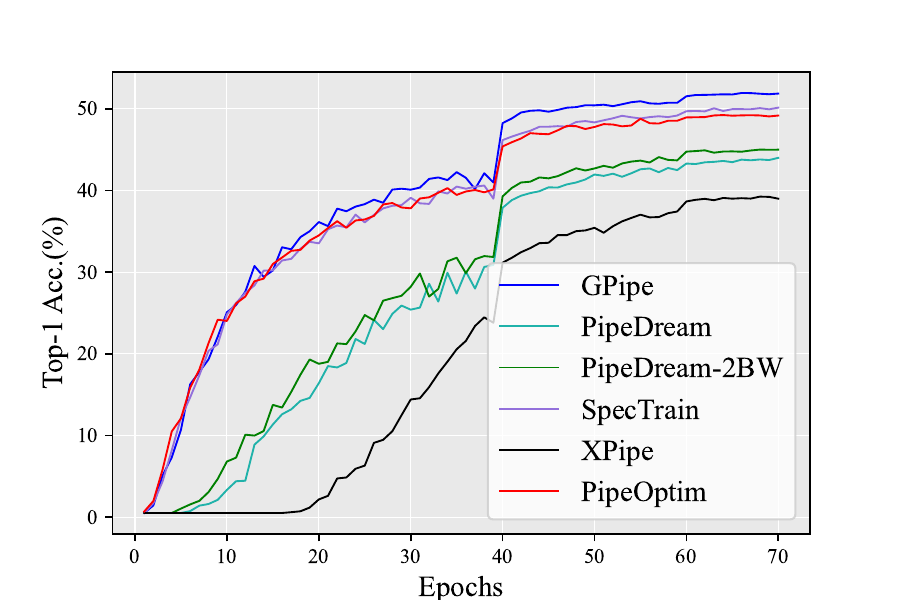}\label{convergence-alexnet-top1}}
				\subfigure[ResNet-101 on Tiny-ImageNet]{\includegraphics[width=0.32\textwidth]{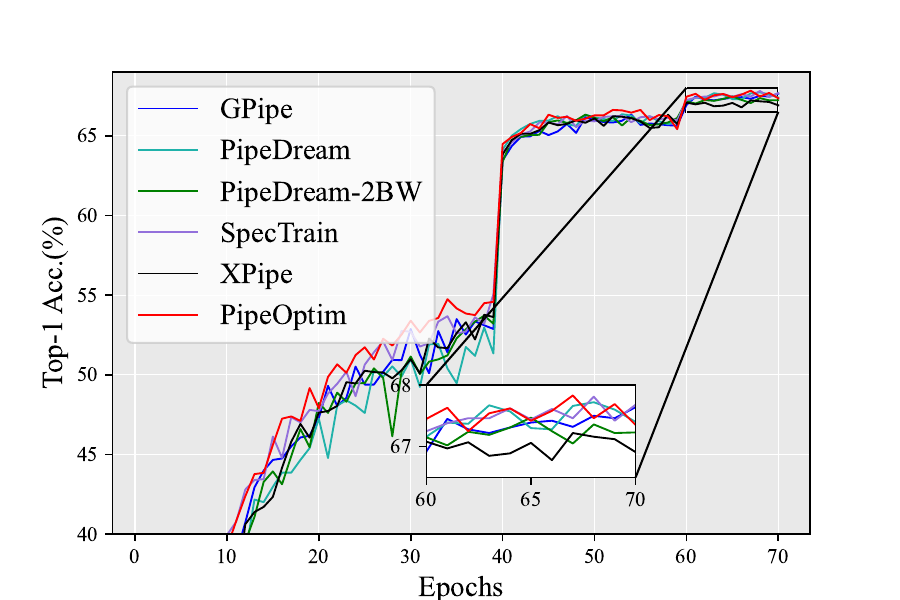}\label{convergence-resnet101-top1}}
				\subfigure[Inception-V3 on Tiny-ImageNet]{\includegraphics[width=0.32\textwidth]{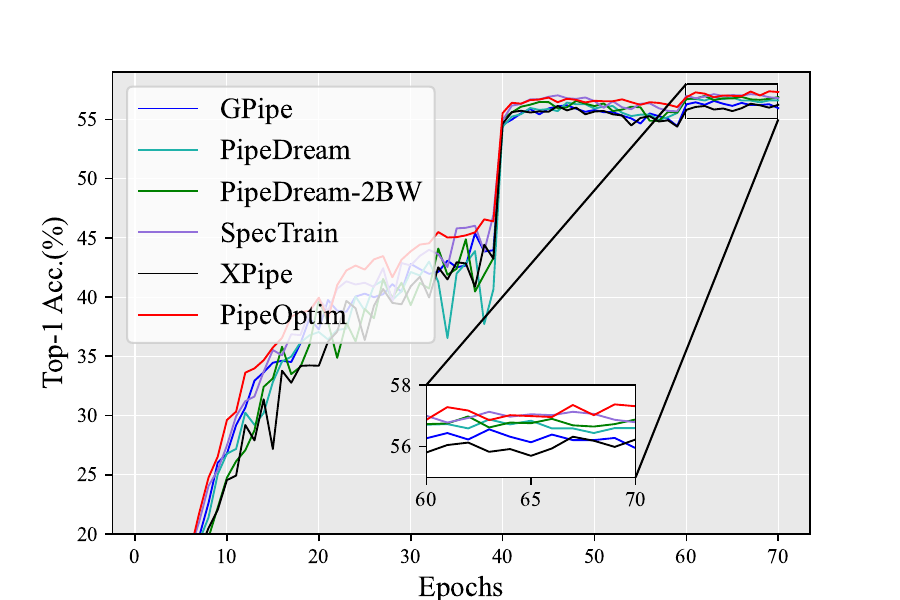}\label{convergence-inceptionv3-top1}}
				\caption{Experiment results of Group-1. Learning curves about top-1 accuracy versus epochs.}
				\label{fig:convergence-group1}
			\end{figure*}
\begin{figure*}[h!]
				\centering
				\subfigure[VGG-16 on CIFAR-100]{\includegraphics[width=0.32\textwidth]{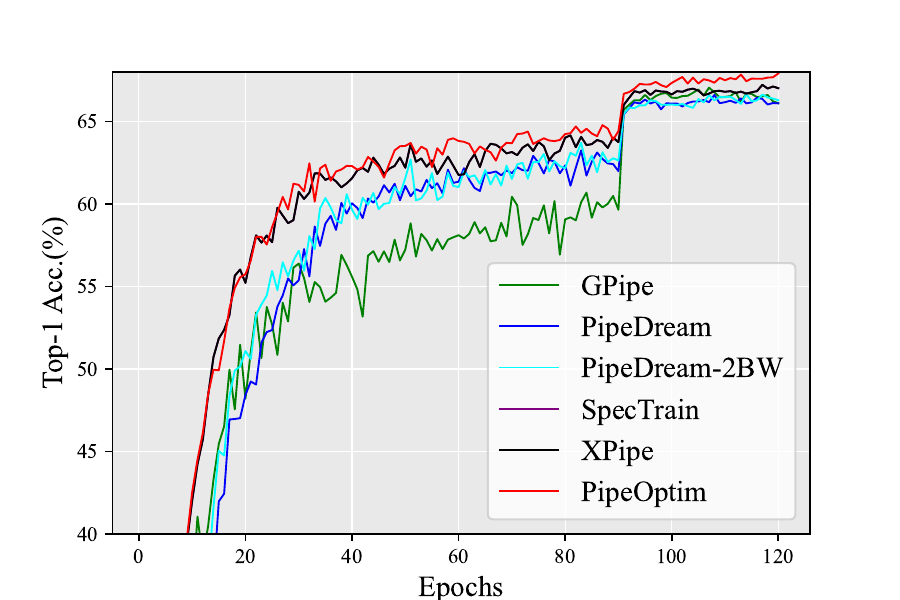}\label{convergence-vgg16-top1}}
				\subfigure[GoogleNet on CIFAR-100]{\includegraphics[width=0.32\textwidth]{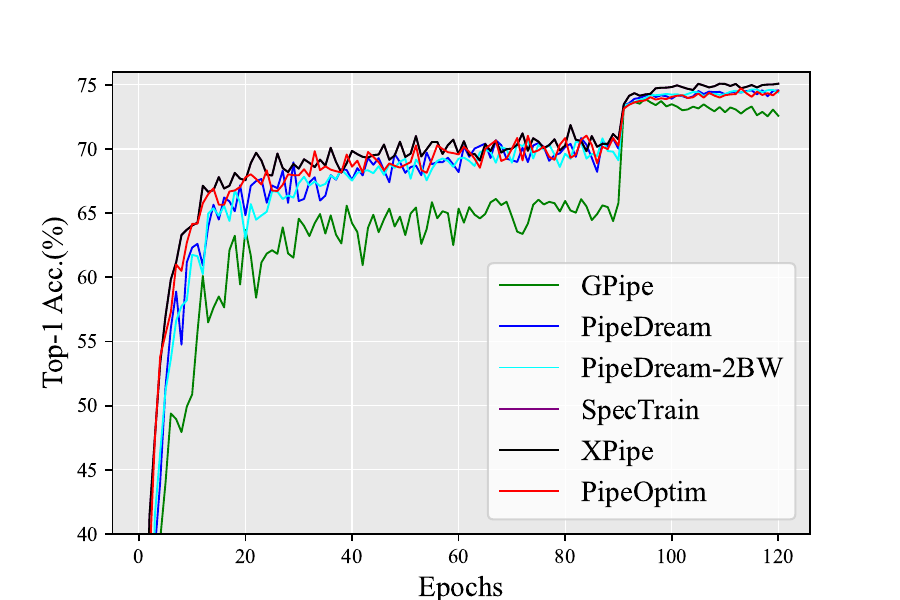}\label{convergence-googlenet-top1}}
				\subfigure[Residual LSTM on IMDB]{\includegraphics[width=0.32\textwidth]{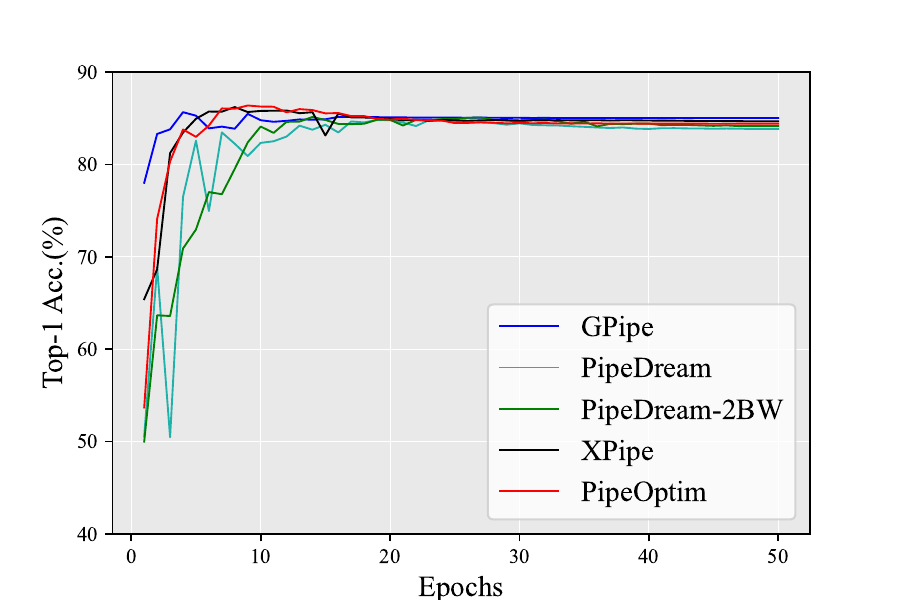}\label{convergence-adamw-lstm-top1}}
				\caption{Experiment results of Group-2.  Learning curves about top-1 accuracy versus epochs.}
				\label{fig:convergence-group2}
			\end{figure*}
\begin{figure*}[h!]
				\centering
				\subfigure[Residual LSTM on IMDb]{\includegraphics[width=0.33\textwidth]{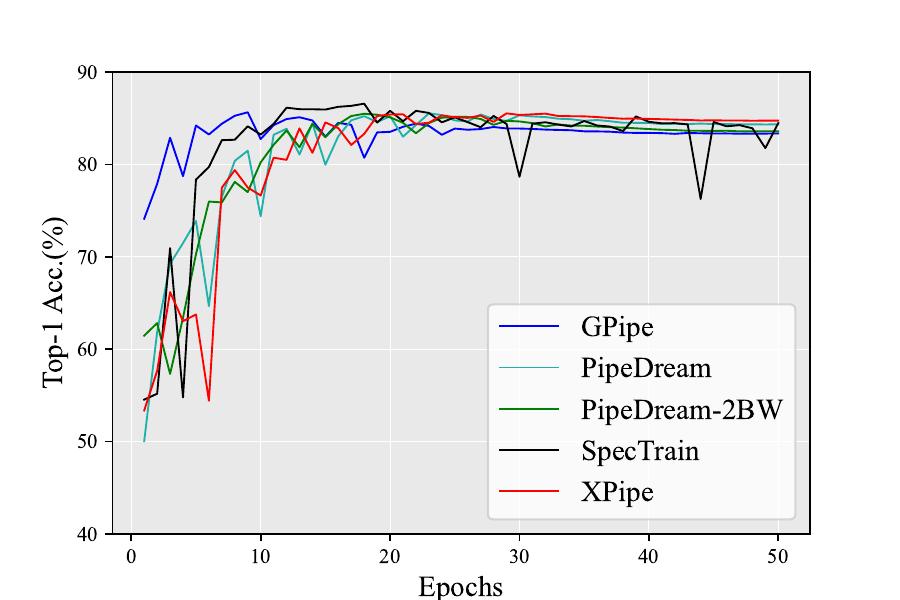}\label{convergence-adam-lstm-top1}}
				\subfigure[GNMT-8 on WMT-16  En$\rightarrow$De]{\includegraphics[width=0.32\textwidth]{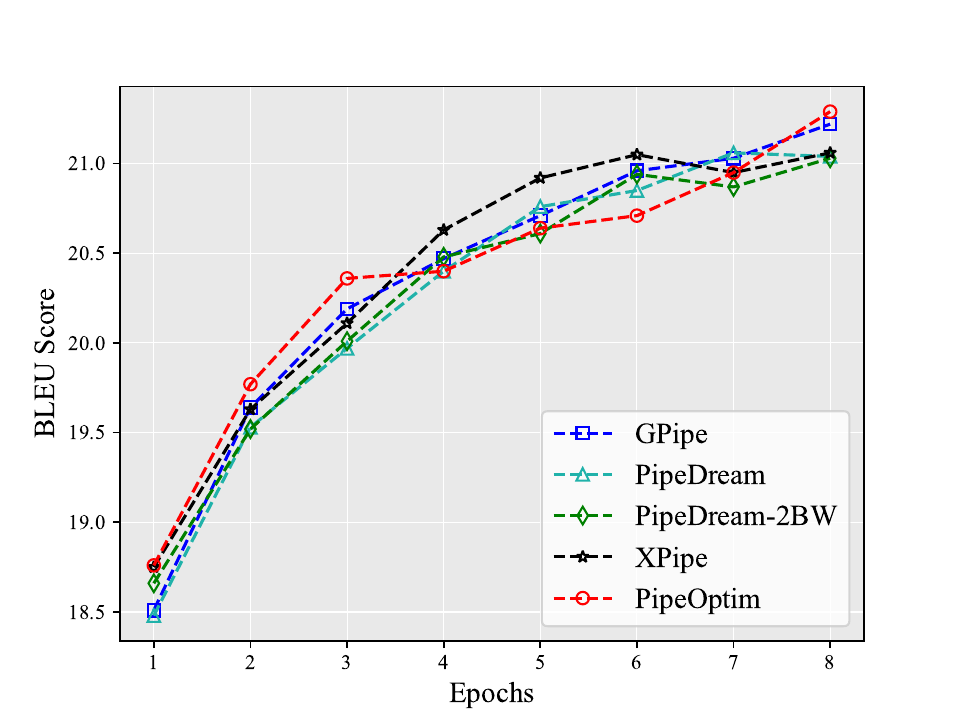}\label{convergence-gnmt8}}
				\subfigure[GNMT-16 on WMT-16  En$\rightarrow$De]{\includegraphics[width=0.32\textwidth]{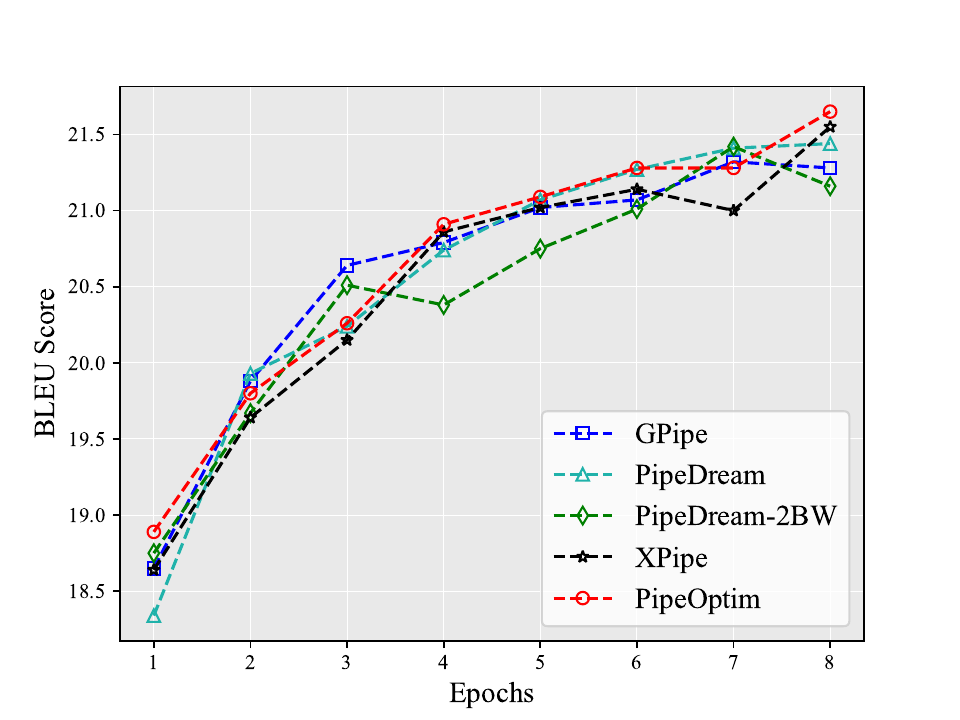}\label{convergence-gnmt16}}
				\caption{Experiment results of Group-3.  Figure~\ref{convergence-adam-lstm-top1}: top-1 accuracy versus epochs; Figures~\ref{convergence-gnmt8} and~\ref{convergence-gnmt16}: BLEU score versus epochs. }
				\label{fig:convergence-group3}
			\end{figure*}

			Figures~\ref{fig:convergence-group1}, \ref{fig:convergence-group2}, and~\ref{fig:convergence-group3} display the learning curves about accuracy versus epochs for Group-1, Group-2, and Group-3, respectively. Table~\ref{tab:acc} summarizes the obtained maximum accuracy of each PMP approach. We can reach the following conclusions based on the experiment results.

		\begin{table*}[h!]
			\caption{Summarization of obtained maximum model accuracy. The best results are highlighted in boldface.}
			\begin{center}
				\begin{tabular}{ccccccccc}
					\toprule
					\multirow{2}{*}{Optimizer} & \multirow{2}{*}{Model}& \multirow{2}{*}{Dataset} & \multicolumn{6}{c}{Approaches} \\
					\cline{4-9}
					&  &  & GPipe & PipeDream &  PipeDream-2BW & SpecTrain & XPipe & PipeOptim \\
					\midrule
					\multirow{3}{*}{\makecell{SGDM}} 
					&  AlexNet & Tiny-ImageNet &\textbf{51.93}\% & 43.99\% & 45.00\% & 50.14\%  & 39.24\% & 49.23\%  \\
					&  ResNet-101 & Tiny-ImageNet & 67.64\% & 67.72\%& 67.47 \%&  67.81\%& 67.22\% &  \textbf{67.84}\%  \\
					& Inception-V3 & Tiny-ImageNet & 56.56\%& 56.88\% &56.98\% & 57.13\% & 56.32\%  & \textbf{57.37}\%\\
					\midrule
					\multirow{3}{*}{\makecell{AdamW}} 
					& Residual LSTM& IMDB& 85.65\% & 84.82\%& 85.13\% &-- & 86.20\% & \textbf{86.38}\%\\
					& VGG-16  & CIFAR-100 & 67.05\% & 66.61\% & 66.66\% & -- & 67.22\% & \textbf{67.91}\% \\
					& GoogleNet & CIFAR-100 & 73.91\%& 74.58\%& 74.64\% & --& \textbf{75.09}\% & 74.73\%\\
					\hline
					\multirow{3}{*}{\makecell{Adam}} 
					& Residual LSTM& IMDB& 85.64\%& 85.52\%& 85.48\%& --& \textbf{86.57}\%& 85.52\%\\
					& GNMT-8  & WMT16 & 
					21.22 BLEU & 21.04 BLEU&  21.03 BLEU &  \textendash & 21.06 BLEU & \textbf{21.29} BLEU  \\
                    & GNMT-16 & WMT16& 21.32 BLEU & 21.44 BLEU & 21.42 BLEU&   \textendash &  21.55 BLEU & \textbf{21.65} BLEU \\
					\bottomrule
				\end{tabular}
				\label{tab:acc}
			\end{center}
		\end{table*}
			
			First, Figures~\ref{convergence-alexnet-top1}, \ref{convergence-resnet101-top1}, and~\ref{convergence-inceptionv3-top1} demonstrate that the learning curves of PipeOptim align closely with those of GPipe when training CNNs using the SGDM as the optimizer. In contrast, PipeDream and PipeDream-2BW consistently show worse convergence compared to both GPipe and PipeOptim. The experiment results in Table~\ref{tab:acc} indicate that GPipe achieves the highest performance when training AlexNet, while PipeOptim obtains the best top-1 accuracy for training ResNet-101 and Inception-V3. Compared to PipeDream and PipeDream-2BW, PipeOptim shows an average improvement of 1.95\% (with a maximum of 5.24\%) and 1.66\% (with a maximum of 4.23\%) in accuracy, respectively. Additionally, PipeOptim demonstrates accuracy comparable to SpecTrain, with both approaches consistently outperforming PipeDream and PipeDream-2BW. These results highlight the superiority of the weight prediction technique over the weight stashing technique in achieving effective parameter learning when using the SGDM to optimize CNN weights. Notably, when SGDM is used as the optimizer, the accuracy obtained by XPipe fluctuates significantly, suggesting that the weight prediction mechanism based on Adam is not always optimal for scenarios where SGDM is employed as the optimizer.
			
			
			
			
			Second, the learning curves shown in Figures~\ref{convergence-vgg16-top1}, \ref{convergence-googlenet-top1}, and~\ref{convergence-adamw-lstm-top1} further confirm the effectiveness of PipeOptim when optimizing the DNN models with the AdamW optimizer. When trained for the same number of epochs, PipeOptim (represented by the red lines) attains higher accuracy than its competitors. Table~\ref{tab:acc} shows that PipeOptim outperforms all other PMP approaches, including the baseline approach, GPipe. Specifically, PipeOptim obtains an average accuracy improvement of 0.80\% (up to 0.86\%) over GPipe.  Compared to PipeDream and PipeDream-2BW, the average accuracy improvements are 1.0\% (up to 1.56\%) and 0.86\% (up to 1.25\%), respectively.  While PipeOptim's accuracy is similar to that of XPipe, it still slightly outperforms XPipe. The experiment results verify that the proposed optimizer-based weight prediction strategy enhances accuracy and, more importantly, is more effective than the weight stashing technique when training the DNN models with the AdamW optimizer.
			
			

			Third, Figure~\ref{convergence-adam-lstm-top1} illustrates the experimental results for training Residual LSTM on the IMDb dataset. Figures~\ref{convergence-gnmt8} and~\ref{convergence-gnmt16} display the experimental results for training GNMT-8 and GNMT-16 on the WMT-16 En$\rightarrow$De dataset. When training Residual LSTM, XPipe achieves the best accuracy, followed by GPipe in second place. PipeOptim and PipeDream are ranked third, and PipeDream-2BW ranks last. For training GNMT-8, PipeOptim converges quickly in the initial stages and ultimately achieves the highest BLEU score (21.29 compared to 21.22 for GPipe, 21.04 for PipeDream, 21.03 for PipeDream-2BW, and 21.06 for XPipe). For training GNMT-16, PipeOptim ultimately achieves a BLEU score of 21.65, again outperforming its competitors. Notably, PipeOptim outperforms both PipeDream and PipeDream-2BW, achieving BLEU scores that are 0.21 and 0.23 higher, respectively. These results demonstrate that PipeOptim ensures effective parameter learning when using the Adam optimizer to train DNN models.
			
			Overall, the experiments demonstrate the effectiveness and robustness of PipeOptim. Notably, the weight prediction scheme proposed by SpecTrain works conditionally, as it is only effective when using the SGDM as the optimizer. XPipe performs well with Adam and AdamW, but its accuracy with SGDM fluctuates significantly depending on the specific deep learning model. In contrast, the performance of PipeOptim is independent of the optimizer type, which is consistently better than that of PipeDream and PipeDream-2BW, and is generally on par or slightly better than GPipe.

	    	\subsection{Throughput}~\label{sec:throughput}
	    	In this subsection, we compare the throughput of PipeOptim with that of GPipe, PipeDream, PipeDream-2BW, SpecTrain, and XPipe. To enable high throughput for GPipe, we evaluated its throughput by splitting each mini-batch into 4 micro-batches, as shown in Figure~\ref{fig:gpipe}. We selected AlexNet, ResNet-101, Inception-V3, VGG-16, GoogleNet, Residual LSTM, GNMT-8, and GNMT-16 as the benchmark neural networks. For GNMT-8 and GNMT-16, the throughput was obtained by training for 2000 iterations. For other DNN models, the maximum throughput was taken after training for 3 epochs. We used the same experimental setup as in Subsection V-B to train these models, except for the mini-batch settings.  Specifically, we always selected an appropriate maximum per-GPU batch size so that each evaluated approach could run normally without yielding out-of-memory (OOM) exceptions. Notably, for fairness, we always let each PMP approach make use of the checkpointing technique~\cite{huang2019gpipe} to reduce the activation memory requirements.



\begin{figure*}[htb]
				\centering
				\subfigure[AlexNet with SGDM]{\includegraphics[width=0.325\textwidth]{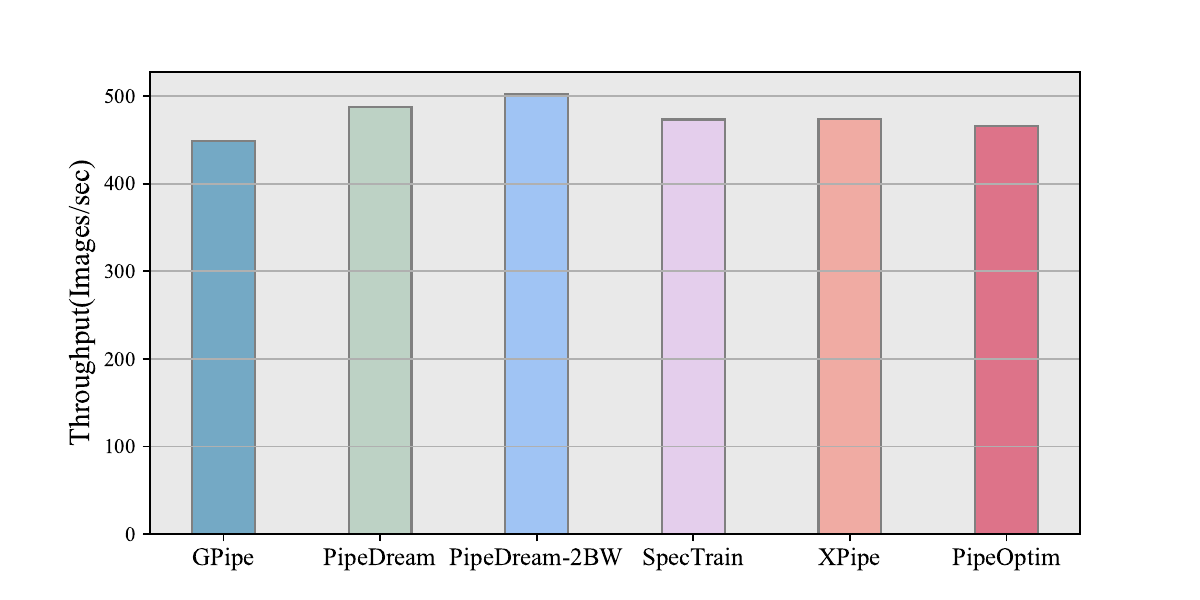}\label{max-throughput-alexnet}}
				\subfigure[ResNet-101 with SGDM]{\includegraphics[width=0.325\textwidth]{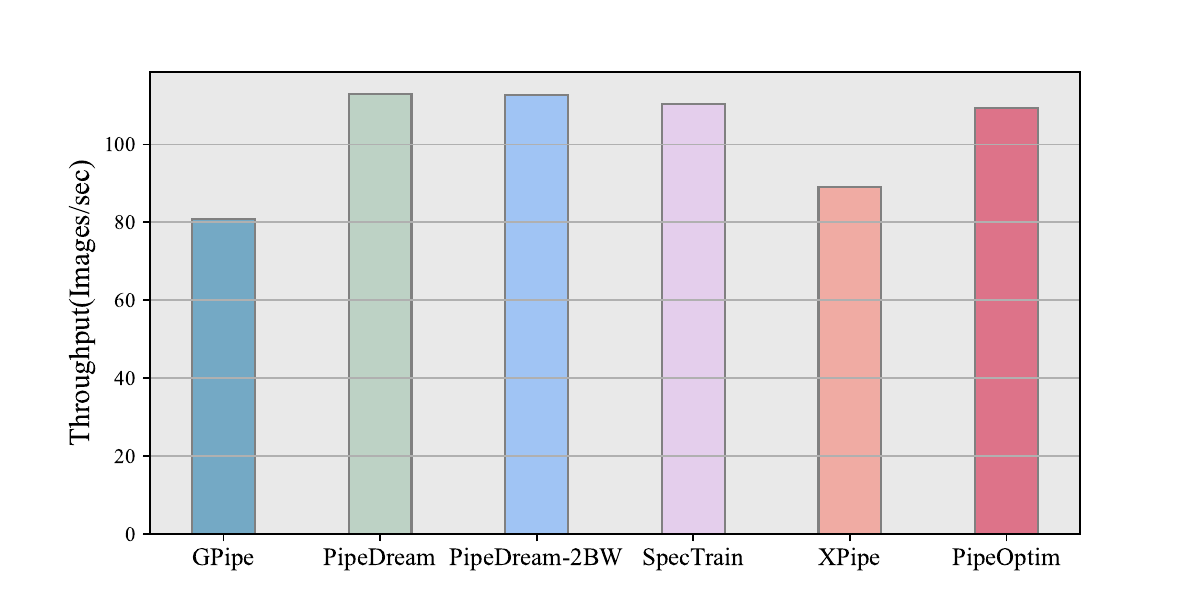}\label{max-throughput-resnet}}
				\subfigure[Inception-V3 with SGDM]{\includegraphics[width=0.325\textwidth]{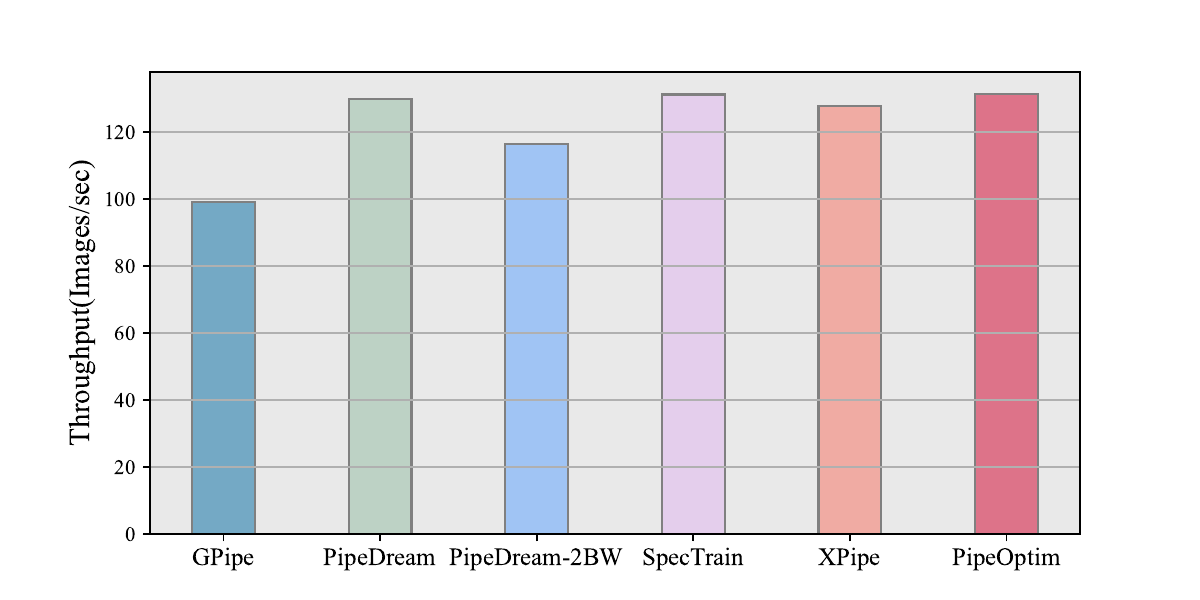}\label{max-throughput-inceptionv3}}
				\subfigure[VGG-16 with AdamW]{\includegraphics[width=0.325\textwidth]{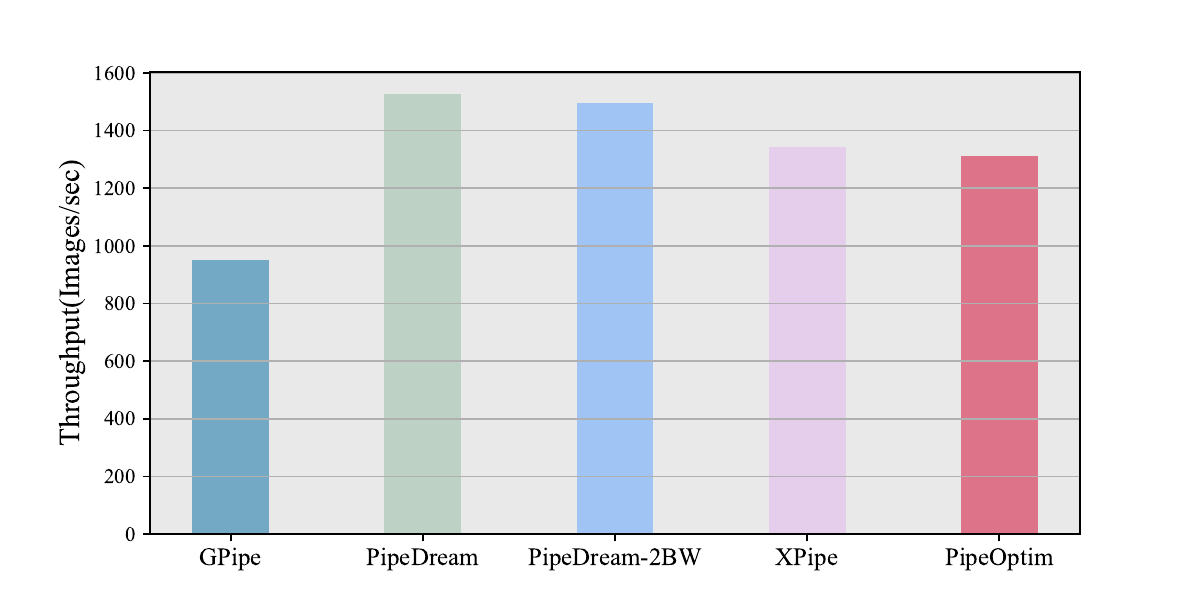}\label{max-throughput-vgg16}}
				\subfigure[GoogleNet with AdamW]{\includegraphics[width=0.325\textwidth]{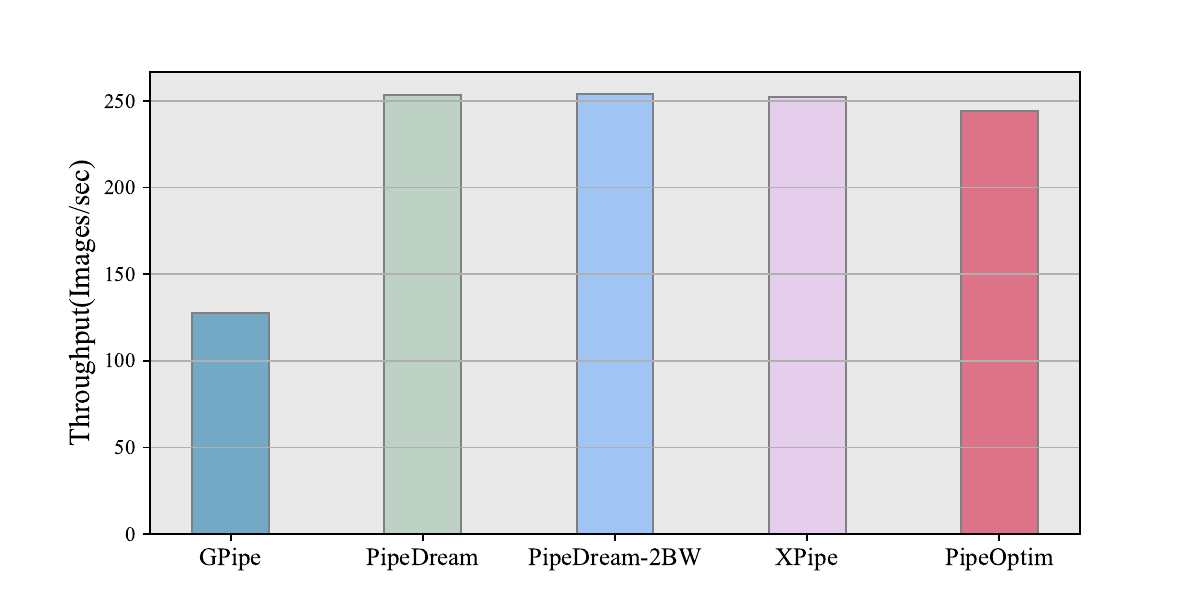}\label{max-throughput-googlenet}}
				\subfigure[Residual LSTM with AdamW]{\includegraphics[width=0.325\textwidth]{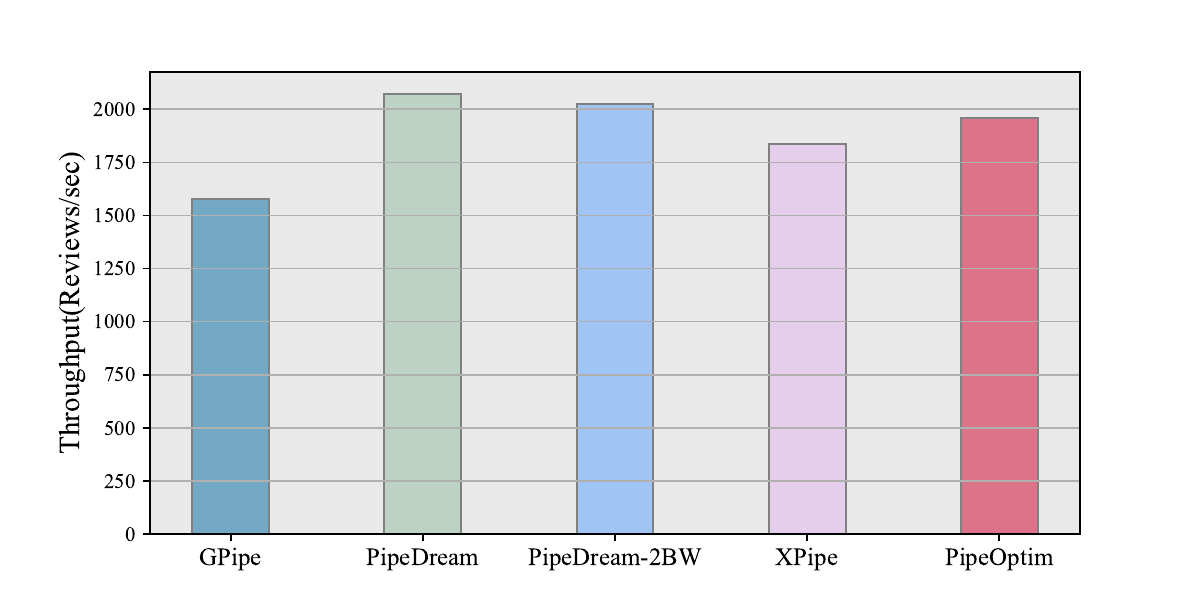}\label{max-throughput-residual}}
				\subfigure[GNMT-8 with Adam]{\includegraphics[width=0.325\textwidth]{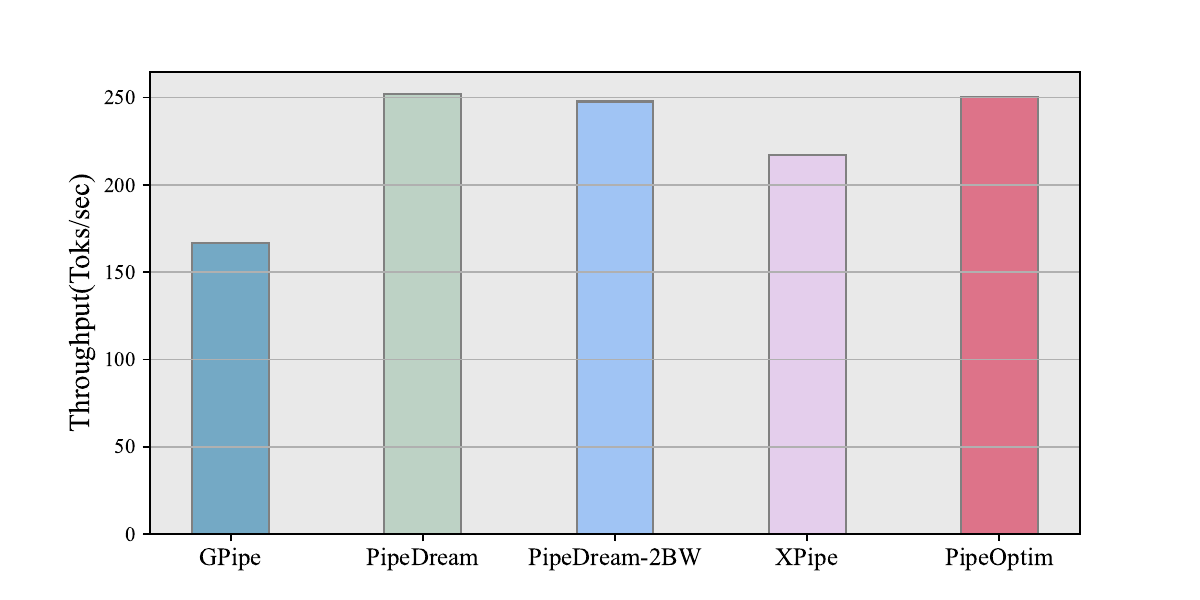}\label{max-throughput-gnmt8}}
				\subfigure[GNMT-16 with Adam]{\includegraphics[width=0.325\textwidth]{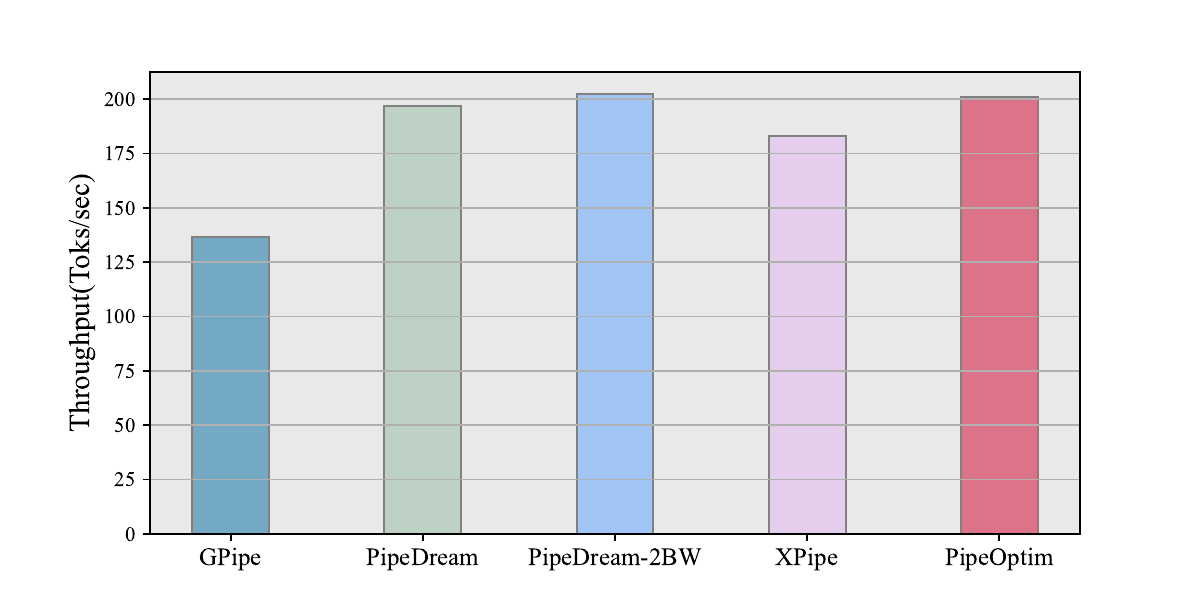}\label{max-throughput-gnmt16}}
				\caption{Throughput results when running with the maximum per-GPU batch size.}
				\label{fig:throughput-B-max}
			\end{figure*}

			Figure~\ref{fig:throughput-B-max} shows the throughput of all the evaluated approaches when training with the maximum per-GPU batch size. Based on the experimental results, we can draw the following conclusions about throughput. First, GPipe obtains the lowest throughput among all evaluated PMP approaches, confirming that GPipe is most affected by bubble overhead. In particular, the experimental results depicted in Figure~\ref{fig:throughput-B-max} show that PipeOptim achieves consistently much higher throughput than GPipe. For example, when training with AdamW optimizer, the throughput of PipeOptim exceeds that of GPipe by 51.27\% (up to 91.7\%). 
			Second, the experimental results show that PipeOptim consistently obtains throughput comparable to that of PipeDream and PipeDream-2BW. The throughput results demonstrate that the throughput of PipeOptim is, on average, 4.0\% (up to 16.5\%) lower than that of PipeDream, and 2.6\% (up to 14.1\%) lower than that of PipeDream-2BW. This demonstrates that the ``1F1B'' strategy is crucial for achieving high throughput in asynchronous PMP approaches. However, weight prediction incurs additional (though not significant) computational overhead compared to weight stashing. Third, note that SpecTrain does not get a throughput results for training VGG-16, GoogleNet, Residual LSTM, GNMT-8 and GNMT-16 because it fails to work with Adam and AdamW optimizers. This highlights the significant limitations of SpecTrain. Fourth, when comparing XPipe and PipeOptim with PipeDream and PipeDream-2BW, it is found that PipeDream and PipeDream-2BW generally achieve higher throughput. This is due to the additional computational overhead caused by performing weight prediction, as opposed to weight stashing. On the other hand, in most cases, the throughput of PipeOptim is always higher than that of XPipe, which confirms that at each iteration, PipeOptim only requires a single weight prediction, resulting in lower computational overhead compared to XPipe, which requires double weight predictions.


			

			\subsection{Memory Consumption} 
			In this subsection, we compare the memory consumption of each PMP approach.  It's well-known that during DNN training, GPU memory is primarily occupied by weights, optimizer states, gradients, activations, and other temporary data~\cite{rajbhandari2020zero}. The key distinction between different PMP approaches lies in the memory consumption for storing weights. Given that GPU storage capacity is limited, higher weight storage requirements lead to a smaller mini-batch size. Therefore, instead of directly reporting the consumed GPU memory, we compare the maximum per-GPU mini-batch size across the evaluated PMP approaches. We selected AlexNet, ResNet-101, Inception-V3, VGG-16, GoogleNet, Residual LSTM, GNMT-8, and GNMT-16 as the evaluated DNN models and assessed them with the same experimental settings as those used by the second strategy in Section~\ref{sec:throughput}. The maximum per-GPU mini-batch sizes are reported in Table~\ref{tab:max-batch-size}.
            
            We can observe that, in most cases, the per-GPU batch size allowed by GPipe is smaller than that of other asynchronous PMP approaches. This is due to the micro-batching technique used by GPipe, which requires storing more intermediate variables on the GPU, thereby consuming more GPU memory. Furthermore, each asynchronous PMP approach typically enables a comparable maximum per-GPU mini-batch size. Specifically, compared to PipeDream, PipeOptim and PipeDream-2BW allow slightly larger mini-batch sizes when training AlexNet, GNMT-8 and GNMT-16. A larger mini-batch size indicates lower memory consumption for storing the weight parameters. This phenomenon validates that PipeOptim and PipeDream-2BW require less GPU memory for storing weights, while PipeDream consumes the most GPU memory for the same purpose. 
			
				\begin{table*}[htb]
				\caption{Summarization of maximum per-GPU mini-batch size.}
				\begin{center}
					\begin{tabular}{ccccccccc}
						\toprule
						Model & Dataset & Optimizer & GPipe &  PipeDream & PipeDream-2BW & SpecTrain & XPipe & PipeOptim \\
						\midrule
						AlexNet & Tiny-ImageNet &SGDM  & 5000/4=1250 & 1240 & 1260 & 1260 & 1250& 1260 \\
						ResNet-101 & Tiny-ImageNet& SGDM& 344/4=86 & 86& 86 & 90 &86 & 86  \\
						Inception-V3 & Tiny-ImageNet & SGDM & 336/4=84  & 84 & 84 & 84 & 84 & 84  \\   
						VGG-16 & CIFAR-100 & AdamW & 8420/4=2105 & 2120& 2120& --& 2800 & 2800 \\
						GoogleNet & CIFAR-100 & AdamW & 920/4=230& 290& 300 &-- & 310 & 295 \\
						Residual LSTM & IMDB & AdamW & 14000/4=3500& 3820 & 3820 &-- & 3820 & 3820 \\
						GNMT-8 &WMT-16 & Adam&  520/4=130 & 136& 140& -- & 142 & 142 \\
						GNMT-16 & WMT-16 &Adam &  496/4=124 & 116& 120& -- & 120 & 120  \\
						\bottomrule
					\end{tabular}
					\label{tab:max-batch-size}
				\end{center}
			\end{table*}

			\subsection{Effectiveness of Weight Prediction}
			In this subsection, we further demonstrate the effectiveness of weight prediction by comparing PipeOptim with the vanilla ``1F1B'' schedule that does not use weight prediction (denoted as Vanilla-1F1B). We selected VGG-16, Inception-V3, and Residual-LSTM as the benchmark models and evaluated them with the same experimental settings described in Subsection~\ref{subsec:exp-convergence}.
			
\begin{figure*}[h!]
				\centering
				\subfigure[VGG-16 on CIFAR-100 with AdamW]{\includegraphics[width=0.32\textwidth]{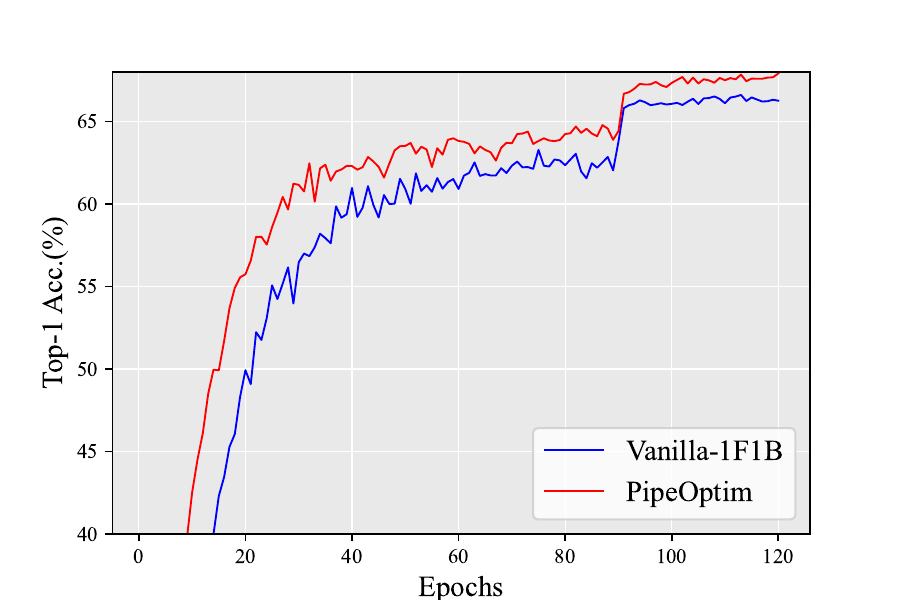}\label{our-1f1b-vgg16-adamw}}
				\subfigure[Inception-V3 on Tiny-ImageNet with SGDM]{\includegraphics[width=0.32\textwidth]{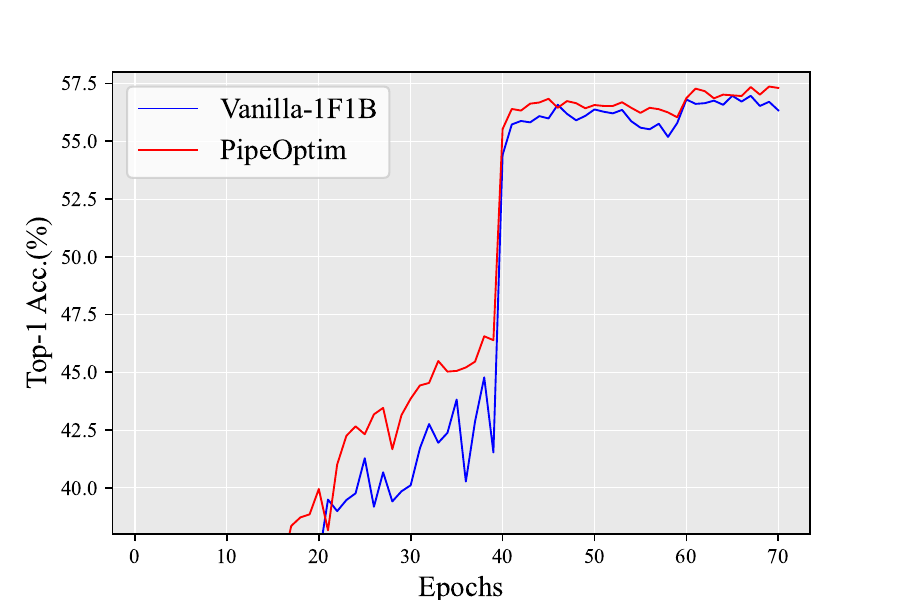}\label{inceptionv-time-accuracy}}
				\subfigure[Residual-LSTM on IMDb with Adam]{\includegraphics[width=0.32\textwidth]{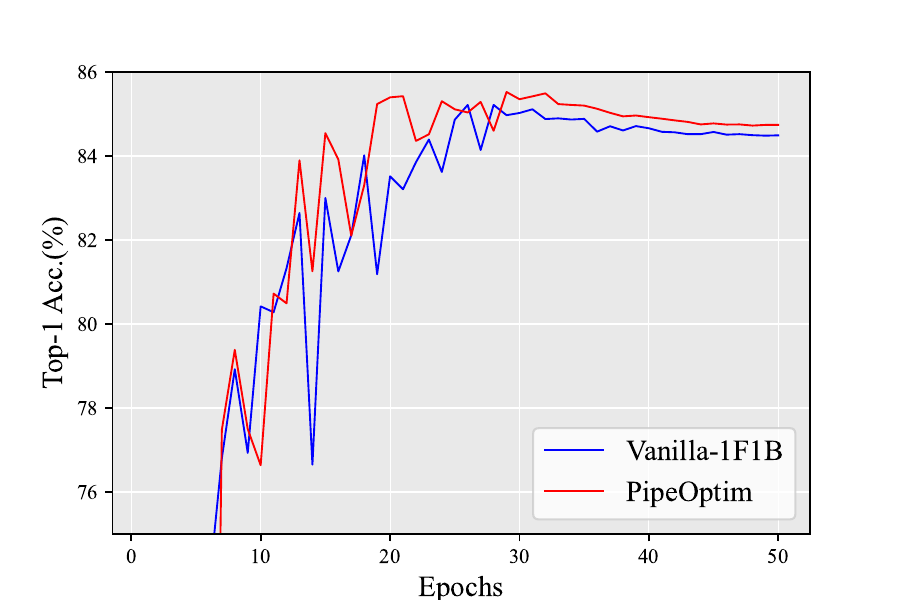}\label{our-1f1b-lstm-adam}}
				\caption{Learning curves about top-1 accuracy versus epochs when comparing PipeOptim with Vanilla-1F1B.}
				\label{fig:our-1f1b}
			\end{figure*}

			 Figure~\ref{fig:our-1f1b} shows the learning curves for top-1 accuracy versus epochs. With the same number of training epochs, PipeOptim consistently achieves higher top-1 accuracy than Vanilla-1F1B, demonstrating significantly better convergence. Additionally, PipeOptim always outperforms Vanilla-1F1B in terms of top-1 accuracy. For instance, when training VGG-16 on CIFAR-100 with the AdamW optimizer, PipeOptim surpasses Vanilla-1F1B by 1.3\%.  On average, PipeOptim achieves 0.67\% higher top-1 accuracy than Vanilla-1F1B. The experimental results strongly validate that the weight prediction strategy contributes to more effective parameter learning. Adding the weight prediction mechanism helps improve the convergence of ``1F1B'', leading to more effective parameter learning.
			 
			
			
			
			
			\subsection{Scalability Study}
			In this subsection, we evaluate the scalability of PipeOptim and compare it with GPipe, PipeDream, PipeDream-2BW, and Vanilla-1F1B. Specifically, we ran the PMP approaches with three different numbers of computing nodes (\#Nodes=1, \#Nodes=2, and \#Nodes=3).  We used Bert-48 as the benchmark model for scalability evaluation and trained it on the Wikipedia dataset. We always set the max sequence length of Bert-48 to 128. Additionally, we employed the same experimental setting as reported in~\cite{li2021chimera}. The Bert-48 model was split into stages, with the number of stages equal to the number of GPUs. In this setup, each GPU is responsible for training a specific stage, maintaining a one-to-one correspondence. 

                \begin{figure*}[htbp]
				\centering
				\includegraphics[width=0.75\textwidth]{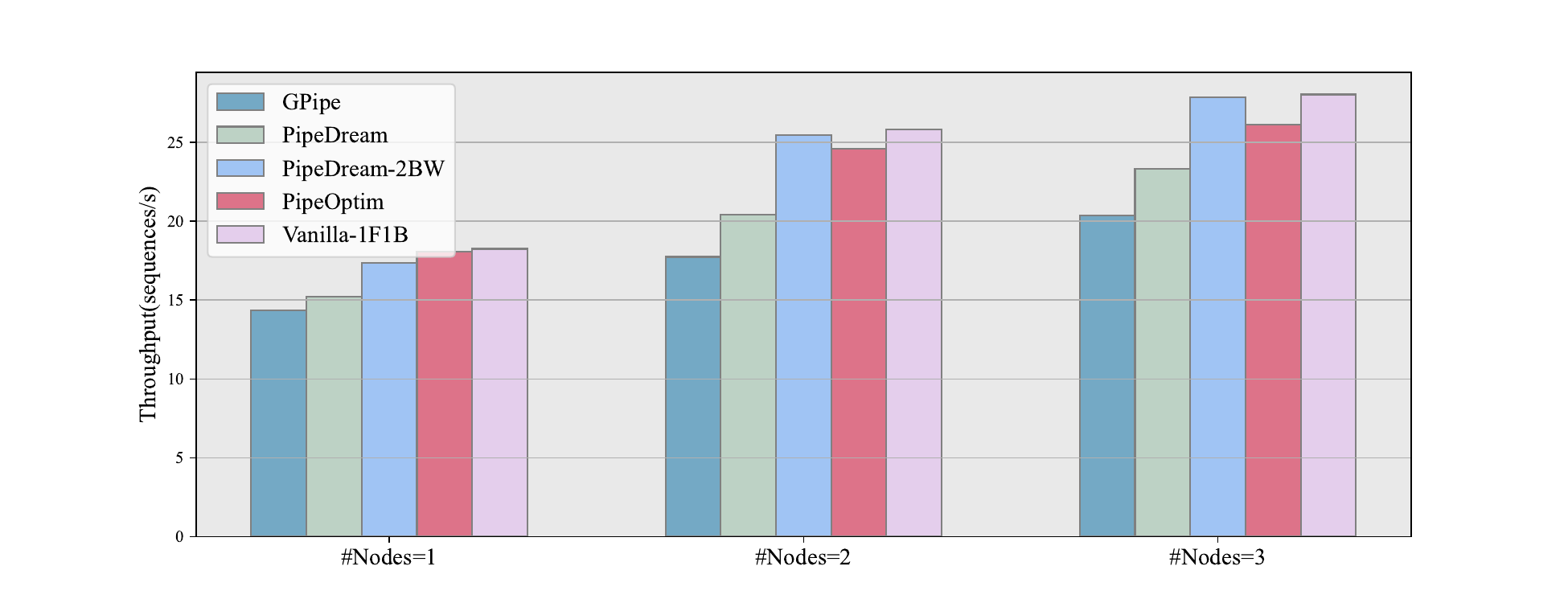}
				\caption{Throughput results when running with different numbers of computing nodes.}
				\label{fig:scalability}
			\end{figure*}
			
			Figure~\ref{fig:scalability} depicts the throughput results when running the PMP approaches with one, two, and three computing nodes. GPipe consistently achieves the lowest throughput, highlighting its inferior scalability compared to the other asynchronous PMP approaches. Among the asynchronous PMP approaches, Vanilla-1F1B demonstrates the best scalability because it does not address the weight inconsistency and weight staleness issues associated with the ``1F1B'' schedule, avoiding the need for additional storage and computation. PipeOptim shows excellent scalability, slightly trailing PipeDream-2BW but outperforming PipeDream. The experimental results demonstrate that PipeOptim can easily scale to multi-machine, multi-GPU environments.
		
				\begin{figure*}[h!]
				\centering
				\subfigure[Top-1 Accuracy vs. Epochs]{\includegraphics[width=0.32\textwidth]{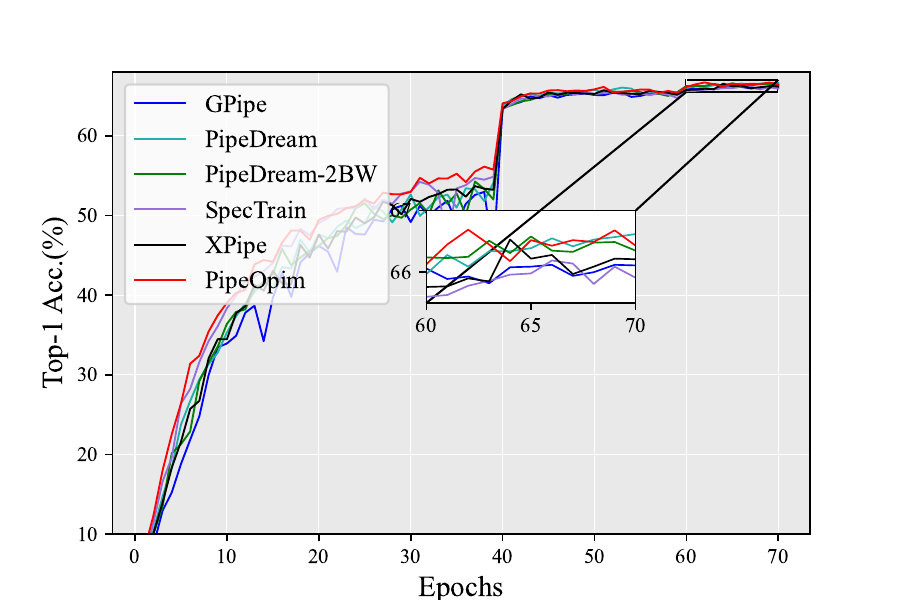}\label{resnet101-top1-epoch}}
				\subfigure[Top-5 Accuracy vs. Epochs]{\includegraphics[width=0.32\textwidth]{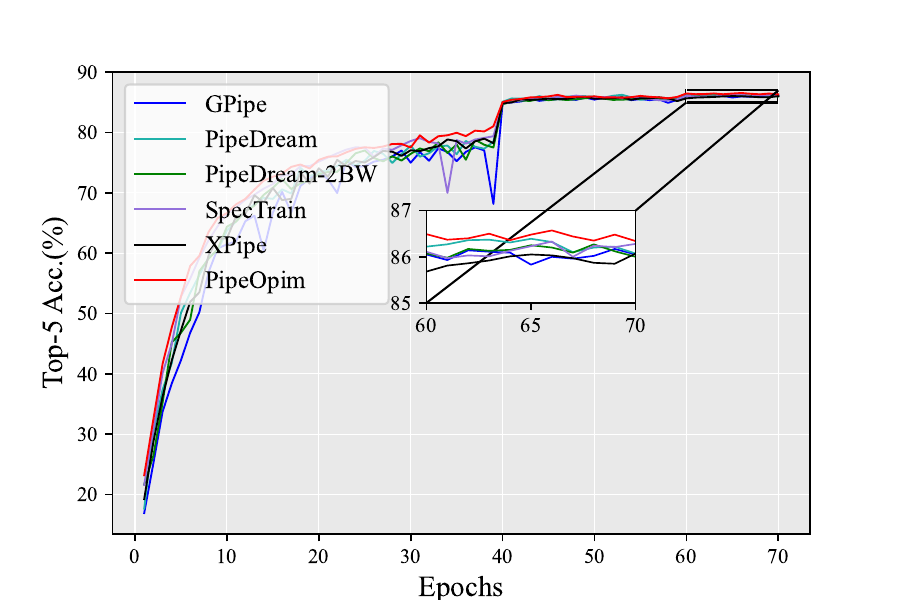}\label{resnet101-top5-epoch}}
				\subfigure[Top-1 Accuracy vs. Time]{\includegraphics[width=0.30\textwidth]{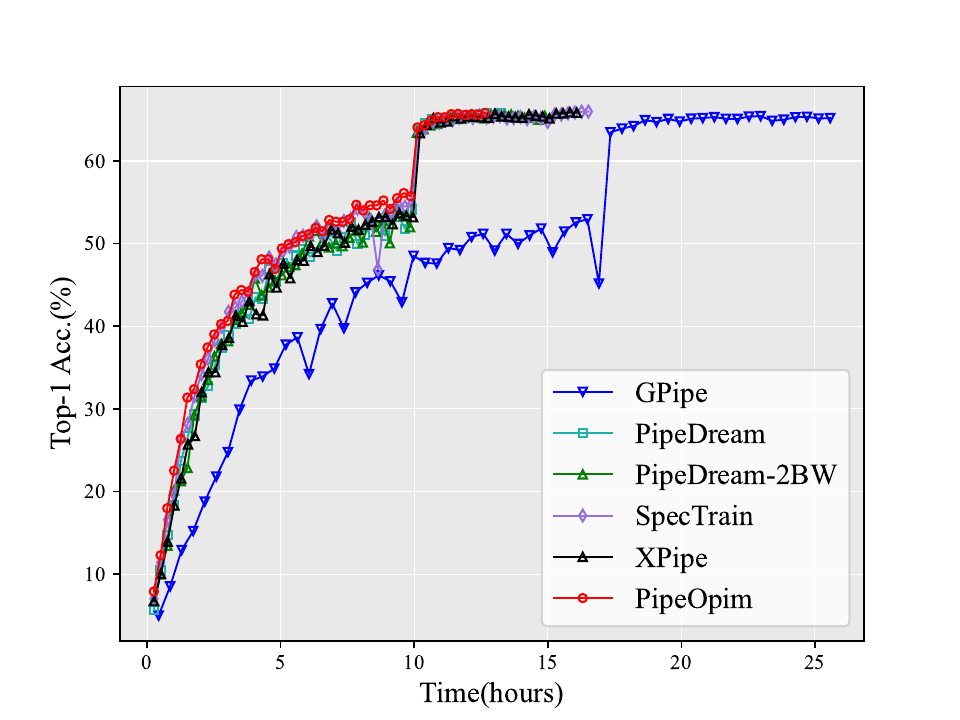}\label{resnet101-time-accuracy}}
				\caption{Overall performance of training ResNet-101 using the SGDM optimizer.}
				\label{fig:overall-performance-resnet101}
			\end{figure*}
			
			\begin{figure*}[h!]
				\centering
				\subfigure[Accuracy vs. Epochs]{\includegraphics[width=0.32\textwidth]{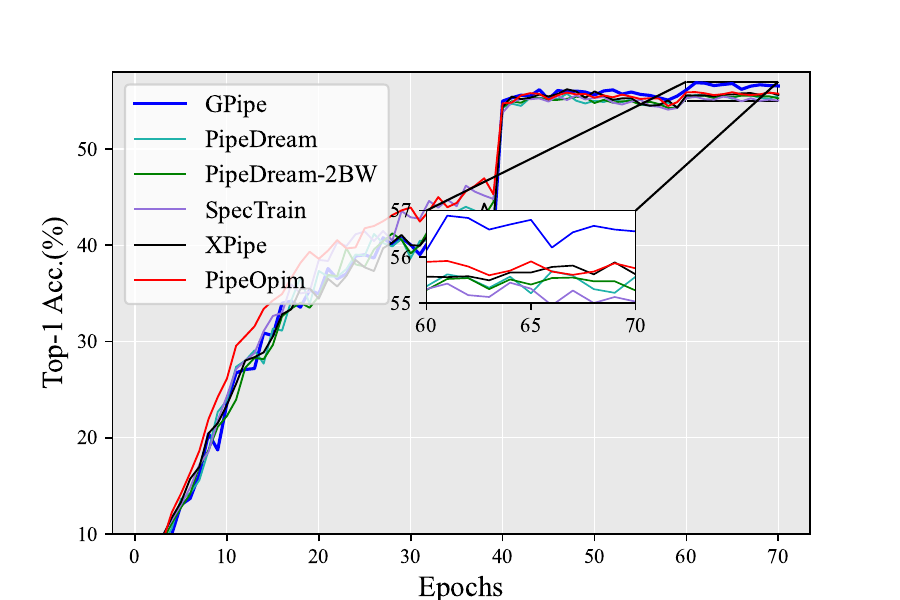}\label{inceptionv3-top1-epoch}}
				\subfigure[Top-5 Accuracy vs. Epochs]{\includegraphics[width=0.32\textwidth]{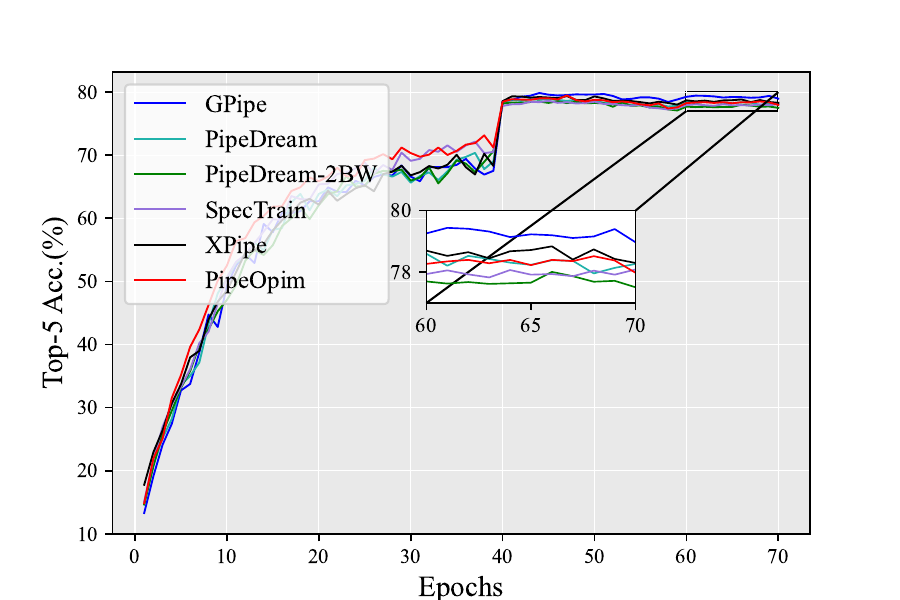}\label{inceptionv3-top5-epoch}}
				\subfigure[Accuracy vs. Time]{\includegraphics[width=0.30\textwidth]{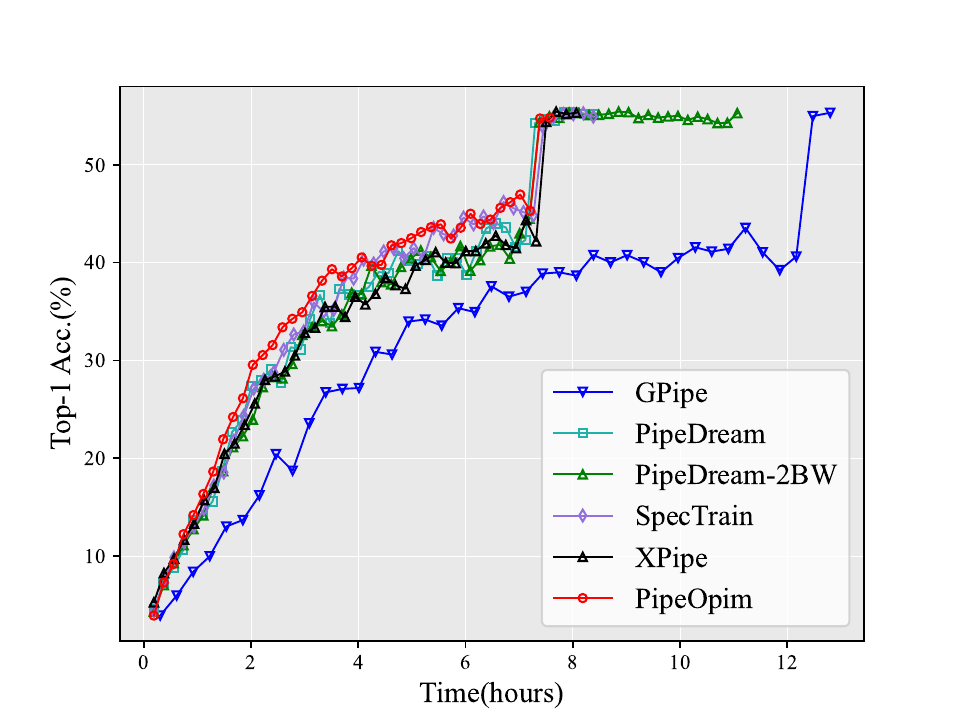}\label{inceptionv3-time-accuracy}}
				\caption{Overall performance of training Inception-V3 using the SGDM optimizer.}
				\label{fig:overall-performance-inceptionv3}
			\end{figure*}

			\subsection{Overall Performance}
			In this subsection, we evaluate the overall performance of all evaluated PMP approaches when simultaneously considering the convergence, throughput, and memory consumption. Specifically, we trained each PMP approach with the maximum per-GPU batch size that allowed each approach to run without triggering OOM exceptions. The pipeline training continued until the target accuracy was reached.  We selected ResNet-101 and Inception-V3 as the benchmark model and conducted the experiments on a computing node. We trained ResNet-101 and Inception-V3 on Tiny-ImageNet using SGDM, following the experimental setup described in Subsection~\ref{subsec:exp-convergence}. Furthermore, we measured the overall performance of each PMP approach by training ResNet-101 and Inception-V3 until the target top-1 accuracies of 66.0\% and 55.5\%, respectively, were achieved.
			
			Figures~\ref{resnet101-top1-epoch}, \ref{resnet101-top5-epoch}, \ref{inceptionv3-top1-epoch}, and~\ref{inceptionv3-top5-epoch} illustrate the learning curves for top-1 and top-5 accuracy versus epochs. Figures~\ref{resnet101-time-accuracy} and~\ref{inceptionv3-time-accuracy} show the relationship between top-1 accuracy and running time until the target accuracy is reached. Table~\ref{tab:overall} summarizes the experiment results on overall performance. We can reach the following conclusions based on the observation of the experiment results. First, PipeOptim consistently achieves higher top-1 and top-5 accuracy than both PipeDream and PipeDream-2BW.  On average, PipeOptim outperforms PipeDream by 0.12\% (up to 0.17\%), PipeDream-2BW by 0.27\% (up to 0.36\%), and SpecTrain by 0.42\% (up to 0.50\%) in top-1 accuracy.
            Second, each asynchronous PMP approach enjoys a comparable 1-epoch running time, but then train faster than GPipe. This indicates that the ``1F1B'' schedule is the primary factor contributing to the training speed of the asynchronous PMP approaches. Third, Pipe  attains comparable accuracy with PipeOptim, but it requires longer iterations times per epoch, and the total time to reach the target accuracy is also longer than PipeOptim.Fourth, thanks to its strong convergence properties, PipeOptim requires the least running time to achieve the target accuracy. Specifically, when training ResNet-101, PipeOptim delivers a speedup of 2.01X over GPipe, 1.04X over PipeDream, 1.17X over PipeDream-2BW, 1.30X over SpecTrain, and 1.27X over XPipe. For Inception-V3, the speedups are 1.69X over GPipe, 1.10X over PipeDream, 1.45X over PipeDream-2BW, 1.10X over SpecTrain, and 1.06X over XPipe.
            
			
			
			
			
			

			
			
			\begin{table}[h!]
				\caption{Results for overall performance evaluation.  In this table, ``1-Epoch Time'' stands for the average running time per epoch, while ``Total Time'' refers to the total training time to reach the target top-1 accuracy (in hours). The best results are highlighted in boldface.}
				\begin{center}
					\begin{tabular}{c|c|c|c}
						\toprule
						Approach & \makecell{Max. Top-1/5 \\ Accuracy} & 1-Epoch Time & \makecell{ Total Time} \\
						\midrule 
						\multicolumn{4}{c}{ResNet-101} \\
						\midrule
						GPipe & 66.12\% / 86.17\% & 1554.07s  & 26.00h   \\ 
						PipeDream & 66.62\% / 86.39\%  & 918.15s & 13.47h  \\ 
						PipeDream-2BW & 66.58\% / 86.27\% & \textbf{911.35}s &  15.14h\\ 
						SpecTrain & 66.19\% / 86.33\% & 914.78s & 16.77h  \\ 
						XPipe & 66.53\% / 86.07\% & 920.21s & 16.34h \\
						PipeOptim & \textbf{66.69}\% / \textbf{86.57}\% & 915.48s & \textbf{12.91}h  \\
						\midrule
						\multicolumn{4}{c}{Inception-V3} \\
						\midrule
						GPipe & \textbf{56.89}\% / \textbf{79.86}\% & 1106.46s   &  13.12h  \\ 
						PipeDream & 55.74\% / 78.91\% & \textbf{656.29}s & 8.57h \\ 
						PipeDream-2BW &  55.55\% / 78.56\% &  664.21s & 11.25h \\ 
						SpecTrain & 55.57\% / 78.63\% & 671.02s & 8.57h \\ 
						XPipe & 56.21\% / 79.37\%& 679.84s & 8.25h \\
						PipeOptim & 55.91\% / 79.48\% & 667.16s & \textbf{7.76}h\\
						\bottomrule
					\end{tabular}
					\label{tab:overall}
				\end{center}
			\end{table}

\section{Related Work}
Pipeline model parallelism has recently been extensively studied to both reduce the communication volume and increase GPU utilization simultaneously~\cite{gaunt2017ampnet, chen2018efficient, ben2018demystifying,huang2019gpipe, narayanan2019pipedream, narayanan2021memory, yang2021pipemare, li2021chimera, he2021pipetransformer, fan2021dapple}. GPipe~\cite{huang2019gpipe} is the most influential synchronous PMP approach. It ensures perfect convergence and incurs no accuracy drop but suffers from serious bubble overhead. Following GPipe, many other synchronous PMP approaches have been proposed, all with a shared goal: reducing bubble overhead by rescheduling mini-/micro-batch arrangements within the pipeline. For example, GEMS~\cite{jain2020gems} and Chimera~\cite{li2021chimera} combine bidirectional pipelines with two versions of weights, training them concurrently in the pipeline. DAPPLE~\cite{fan2021dapple} introduced an early backward scheduling strategy, where the backward tasks were scheduled earlier, thus freeing up memory used for storing activations generated by corresponding forward tasks. To improve efficiency with a single version of weights, Hanayo~\cite{liu2023hanayo} proposes dynamically changing the pipeline direction during the computation process, transforming the pipeline into a wavy-shaped pipeline. Remarkably, an alternative approach to reducing the bubble overhead is to fill them with computations, which can also enhance convergence. One such example is Pipefisher~\cite{osawa2023pipefisher}, which aims to reduce bubbles by introducing additional work during idle periods. Recently, Zero Bubble Pipeline Parallelism (ZB)~\cite{qi2024zero} has been proposed, which nearly eliminates bubble overhead. ZB splits the backward computation into two parts: gradient computation for the input and computation for the parameters, filling the bubbles with useful computations. ZB achieves high GPU utilization on the premise of splitting the backward propagation and further dividing a mini-batch into a a large number of micro-batches. This approach fundamentally differs from asynchronous PMP approaches (such as PipeDream and PipeOptim), which rely entirely on the ``1F1B'' schedule.

For asynchronous pipeline training, the ``1F1B'' schedule is widely used in popular asynchronous PMP approaches, including PipeDream~\cite{narayanan2019pipedream}, PipeDream-2BW~\cite{narayanan2021memory},  WPipe~\cite{yang2021group}, SpecTrain~\cite{chen2018efficient}, and XPipe~\cite{guan2019xpipe}. PipeDream~\cite{narayanan2019pipedream} adopts the weight stashing technique to address the weight inconsistency issue, while PipeDream-2BW~\cite{narayanan2021memory} utilizes the double-buffered weight updates (2BW) technique to reduce the memory consumption of PipeDream. WPipe~\cite{yang2021group} further makes use of a technique called double-grouped weight updates (2GW) to outperform PipeDream-2BW in terms of memory efficiency and fresh weight updates. In contrast to PipeDream, PipeDream-2BW, and WPipe, SpecTrain~\cite{chen2018efficient} adopts SGDM-based weight prediction to enable effective parameter learning when using SGDM as the optimizer. XPipe~\cite{guan2019xpipe} always constructs its weight prediction formula based on the update rule of Adam, which does not depend on the type of optimizer used. Furthermore, PipeMare~\cite{yang2021pipemare} improves the statistical efficiency of asynchronous pipeline parallelism by using the learning rate rescheduling and discrepancy correction. AvgPipe~\cite{chen2023elastic} proposes using the elastic averaging technique~\cite{zhang2015deep} to maintain the statistical efficiency of the execution of multiple pipelines.
			
PMP can be combined with data parallelism to scale pipeline parallelism from single-node systems to multi-node, multi-GPU systems. Many popular PMP approaches, such as PipeDream~\cite{narayanan2019pipedream}, PipeDream-2BW~\cite{narayanan2021memory}, Chimera~\cite{li2021chimera}, and DAPPLE~\cite{fan2021dapple}, support this hybrid-parallel training manner. An alternative form of hybrid parallelism involves simultaneously integrating data parallelism, tensor parallelism, and PMP (known as 3D parallelism). Examples of this approach include DistBelief~\cite{dean2012large}, DeepSpeed, and Megatron-LM~\cite{narayanan2021efficient}, all of which are specifically designed for training large-scale deep neural network models.
			
\section{Discussions}
Adopting the ``1F1B'' scheduling can ensure high GPU utilization, leading to high throughput, but it inevitably results in issues of weight inconsistency and weight staleness. For asynchronous PMP approaches, effective parameter learning and high throughput are equally important, as the total time to train a model is determined by both convergence and the iteration time per epoch. PipeDream and PipeDream-2BW use weight stashing technique to ensure weight consistency, but they do not address the problem of weight staleness. SpecTrain only works when using SGDM as the optimizer, which presents significant limitations. XPipe constructs its weight prediction formula based on the parameter update rules of Adam. However, as demonstrated in the experiments, when using non-Adam optimizers (such as SGDM) to optimize DNN models, effective parameter learning cannot always be guaranteed. In contrast, PipeOptim dynamically constructs weight update formulas based on the update rules of the used optimizer, which can simultaneously address the issues of weight inconsistency and weight staleness. This approach achieves a good trade-off among GPU utilization, effective parameter learning, and memory consumption. As a result, PipeOptim delivers the best overall performance among all evaluated PMP approaches. As validated in Subsection V-G, PipeOptim consistently requires the least training time to achieve the target accuracy for DNN models.

Despite PipeOptim's superiority over five other popular PMP approaches, it still has the following limitations. First, PipeOptim requires each GPU to maintain up to two versions of weights. Although this is less memory-intensive than PipeDream, it may still be a consideration in terms of memory usage, especially when GPU resources are limited. Second, while PipeOptim largely resolves the weight inconsistency and staleness issues caused by the ``1F1B'' schedule, it, like other asynchronous PMP approaches, cannot guarantee the exact same semantics as synchronous PMP approaches. Third, the weight prediction scheme is based on the observation that the update values calculated by each gradient-based optimizer should reflect the ``correct'' direction for updating the weights~\cite{guan2023weight,guan2024xgrad}. However, from a theoretical perspective, the degree to which weight prediction aligns with the true weights cannot be effectively measured. In summary, despite PipeOptim's significant progress in improving GPU utilization and ensuring effective parameter learning, there is still considerable room for improvement.


\section{Conclusions and Future Work}
In this work, we propose an efficient asynchronous PMP approach called PipeOptim. The key insight behind PipeOptim is its use of an optimizer-dependent weight prediction strategy, which simultaneously address the weight inconsistency and weight staleness issues caused by the ``1F1B'' schedule. This approach leads to three major innovations. First, PipeOptim effectively handles the staleness issue that is left unsolved by the weight stashing technique in PipeDream and PipeDream-2BW, resulting in better convergence and higher accuracy. Second, PipeOptim's efficiency is independent of the optimizer used, contrasting sharply with SpecTrain, which works well only when using the SGDM as the optimizer. Third, PipeOptim requires each GPU to maintain at most two versions of the weights. Extensive evaluations across four machine-learning tasks validate the effectiveness of PipeOptim. Our research introduce a new way for efficient and effective asynchronous pipeline training.

Finally, we outline the future directions of our work as follows. First, we plan to conduct a deeper investigation on the mechanism of weight prediction and explore a more effective way to measure how well the weight prediction method aligns with the true weights. Second, building on PipeOptim, we aim to integrate the tensor model parallelism (TMP) training mode and explore ways to accelerate the training of large language models. Third, we will develop a dynamic GPU memory balancing mechanism to enable more efficient memory distribution across GPUs during pipeline training with PipeOptim, further enhancing the computational power of multi-GPU systems. Lastly, pipeline parallelism has already been widely applied to the fine-tuning and training large language models~\cite{eliad2021fine}. In our future research, we plan to incorporate large model training and fine-tuning tasks into our experimental evaluations while continuing to improve the performance of PipeOptim.



			\section*{Acknowledgment}
			Lei Guan thanks Prof. Shigang Li at Beijing University of Posts and Telecommunications (BUPT) for stimulating discussions about distributed training of PMP approaches in multi-machine, multi-GPU settings. The authors thank all the reviewers for their beneficial comments.

			\ifCLASSOPTIONcaptionsoff
			\newpage
			\fi

			\bibliographystyle{IEEEtran}
			\bibliography{mybib}
			
			
			
			%
			
			\begin{IEEEbiography}[{\includegraphics[width=1in,height=1.25in,clip,keepaspectratio]{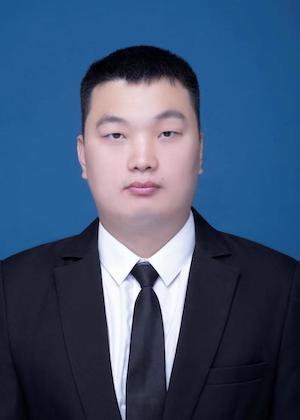}}]{Lei Guan}
				received the Ph.D. degree in Computer Science and Technology from the National University of Defense Technology (NUDT) in 2022. He is currently an associate professor with the College of Computer Science and Technology, Taiyuan University of Technology. His research interests include deep learning, parallel and distributed computing, and optimization. 
			\end{IEEEbiography}
			
			\begin{IEEEbiography}[{\includegraphics[width=1in,height=1.25in,clip,keepaspectratio]{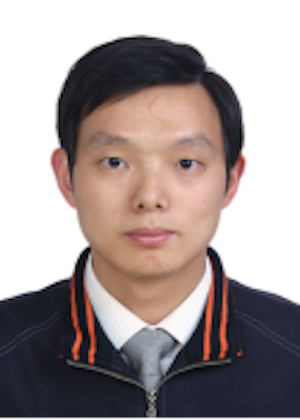}}]{Dongsheng Li}
				received the Ph.D. degree in Computer Science and Technology from the National University of Defense Technology (NUDT) in 2005. He is a professor and doctoral supervisor in the College of Computer at NUDT. He was awarded the Chinese National Excellent Doctoral Dissertation in 2008. His research interests include distributed systems, cloud computing, and big data processing.
			\end{IEEEbiography}

			\begin{IEEEbiography}[{\includegraphics[width=1in,height=1.25in,clip,keepaspectratio]{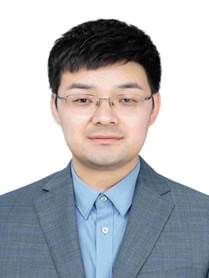}}]{Yongle Chen} was born in Weifang, Shandong, China, in 1983. He received the B.S. degree in computer science from Jilin University in 2007, the M.S. and Ph.D. degrees in computer science from the Institute of Software, Chinese Academy of Science in 2009 and 2013, respectively. He is currently a Full Professor with the College of Computer Science and Technology, Taiyuan University of Technology, Taiyuan, China. His research interests include wireless network, cyber-physical systems, and IoT security.
			\end{IEEEbiography}

			\begin{IEEEbiography}[{\includegraphics[width=1in,height=1.25in,clip,keepaspectratio]{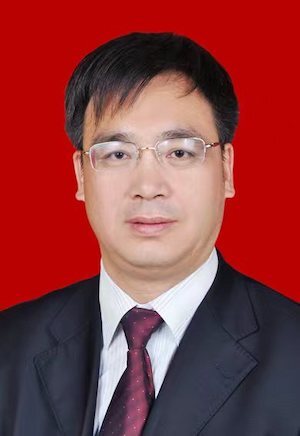}}]{Jiye Liang}
				received the Ph.D. degree from Xi’an Jiaotong University. He is a professor with the Key Laboratory of Computational Intelligence and Chinese Information Processing of the Ministry of Education, School of Computer and Information Technology, Shanxi University. His research interests include artificial intelligence, granular computing, data mining, and machine learning. He has published more than 300 papers in his research fields, including the Artificial Intelligence, Journal of Machine Learning Research, IEEE Transactions on Pattern Analysis and Machine Intelligence, IEEE Transactions on Knowledge and Data Engineering, Machine Learning, NIPS, ICML, and AAAI.
			\end{IEEEbiography}
		
		\begin{IEEEbiography}[{\includegraphics[width=1in,height=1.25in,clip,keepaspectratio]{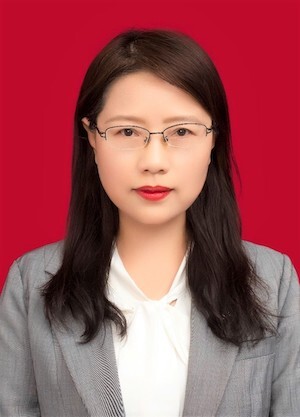}}]{Wenjian Wang}
			received the Ph.D. degree in applied mathematics from Xi'an Jiaotong University, China, in 2004. Now she is a full professor and Ph.D. supervisor of the Key Laboratory of Computational Intelligence and Chinese Information Processing of the Ministry of Education, Shanxi University. She has published more than 270 academic papers, and her research interests include machine learning, data mining, and intelligent computing, etc.
		\end{IEEEbiography}
		
		\begin{IEEEbiography}[{\includegraphics[width=1in,height=1.25in,clip,keepaspectratio]{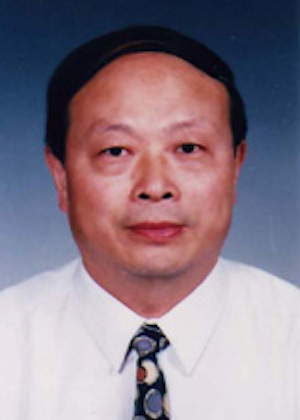}}]{Xicheng Lu}
		received the B.S. degree in computer science from the Harbin Military Engineering Institute, Harbin, China, in 1970. He was a Visiting Scholar with the University of Massachusetts from 1982 to 1984. He is currently a Professor with the College of Computer, National University of Defense Technology, Changsha, Hunan, China. His research interests include distributed computing, computer networks, and parallel computing. He has served as a member of editorial boards of several journals and has co-chaired many professional conferences. He was a joint recipient of more than a dozen academic awards, including four First Class National Scientific and Technological Progress Prize of China. He is an Academician of the Chinese Academy of Engineering.
		\end{IEEEbiography}
		
		\end{document}